\documentclass[10pt,journal,compsoc]{IEEEtran}
\ifCLASSOPTIONcompsoc
  \usepackage[nocompress]{cite}
\else
  \usepackage{cite}
\fi
\ifCLASSINFOpdf
  \usepackage[pdftex]{graphicx}
  \graphicspath{{../pdf/}{../jpeg/}}
  \DeclareGraphicsExtensions{.pdf,.jpeg,.png}
\else
\fi
\usepackage{amsmath}
\usepackage{algorithmic}

\usepackage{array}

\ifCLASSOPTIONcompsoc
  \usepackage[caption=false,font=footnotesize,labelfont=sf,textfont=sf]{subfig}
\else
  \usepackage[caption=false,font=footnotesize]{subfig}
\fi
\usepackage{url}

\usepackage{amsmath}
\usepackage{amssymb}
\usepackage{multirow}
\usepackage{booktabs}
\usepackage{array}
\usepackage{bm}
\usepackage{bbm}
\usepackage[table]{xcolor}
\usepackage{placeins}

\usepackage[flushleft]{threeparttable}

\usepackage[colorlinks,hyperindex,breaklinks]{hyperref}
\usepackage[misc,geometry]{ifsym}

\begin{document}
%
\title{Meta-DETR: Image-Level Few-Shot Detection \\ with Inter-Class Correlation Exploitation}
%
%
%
%

\author{Gongjie Zhang$^\dagger$,
        Zhipeng Luo$^\dagger$,
        Kaiwen Cui,
        Shijian Lu\,$^{\text{\Letter}}$,
        and~Eric P. Xing
\IEEEcompsocitemizethanks{\IEEEcompsocthanksitem Gongjie Zhang, Zhipeng Luo, Kaiwen Cui, and Shijian Lu are with the School of Computer Science and Engineering, Nanyang Technological University, Singapore.\protect
\IEEEcompsocthanksitem Eric P. Xing is with the School of Computer Science, Carnegie Mellon University. He also serves as the president of Mohamed bin Zayed University of Artificial Intelligence.\protect
\IEEEcompsocthanksitem E-mail: GongjieZhang@ntu.edu.sg, Zhipeng001@e.ntu.edu.sg, Kaiwen00\\1@e.ntu.edu.sg, Shijian.Lu@ntu.edu.sg, Eric.Xing@mbzuai.ac.ae.
\IEEEcompsocthanksitem $\dagger$ denotes equal contribution; \;${\text{\Letter}}$ denotes corresponding author.}
\thanks{Pre-print version. All rights reserved by the authors.}}

%
%

\markboth{ZHANG \textit{ET AL.}: Meta-DETR: Image-Level Few-Shot Detection with Inter-Class Correlation Exploitation}%
{Shell \MakeLowercase{\textit{et al.}}: Bare Demo of IEEEtran.cls for Computer Society Journals}
%



\IEEEtitleabstractindextext{%
\begin{abstract}

Few-shot object detection has been extensively investigated by incorporating meta-learning into region-based detection frameworks. Despite its success, the said paradigm is still constrained by several factors, such as \textit{(i)} low-quality region proposals for novel classes and \textit{(ii)} negligence of the inter-class correlation among different classes. Such limitations hinder the generalization of base-class knowledge for the detection of novel-class objects. In this work, we design Meta-DETR, which \textit{(i)} is the first image-level few-shot detector, and \textit{(ii)} introduces a novel inter-class correlational meta-learning strategy to capture and leverage the correlation among different classes for robust and accurate few-shot object detection. Meta-DETR works entirely at image level without any region proposals, which circumvents the constraint of inaccurate proposals in prevalent few-shot detection frameworks. In addition, the introduced correlational meta-learning enables Meta-DETR to simultaneously attend to multiple support classes within a single feedforward, which allows to capture the inter-class correlation among different classes, thus significantly reducing the misclassification over similar classes and enhancing knowledge generalization to novel classes. Experiments over multiple few-shot object detection benchmarks show that the proposed Meta-DETR outperforms state-of-the-art methods by large margins. The implementation codes are available at \url{https://github.com/ZhangGongjie/Meta-DETR}.

\end{abstract}

}

\maketitle

\IEEEdisplaynontitleabstractindextext

\ifCLASSOPTIONpeerreview
\begin{center} \bfseries EDICS Category: 3-BBND \end{center}
\fi
%
\IEEEpeerreviewmaketitle

\IEEEraisesectionheading{\section{Introduction}\label{sec:introduction}}

\IEEEPARstart{C}{omputer} vision has experienced significant progress in recent years. However, there still exists a huge gap between current computer vision techniques and the human visual system in learning new concepts from very few examples: most existing methods require a large amount of annotated samples, while humans can effortlessly recognize a new concept even with just a glimpse of it~\cite{Landau1988TheIO}. Such human-like capability to generalize from limited examples is highly desirable for machine vision systems, especially when sufficient training samples are unavailable or their annotations are hard to obtain.

\begin{figure}[t!] 
\begin{center}
   \includegraphics[width=0.95\linewidth]{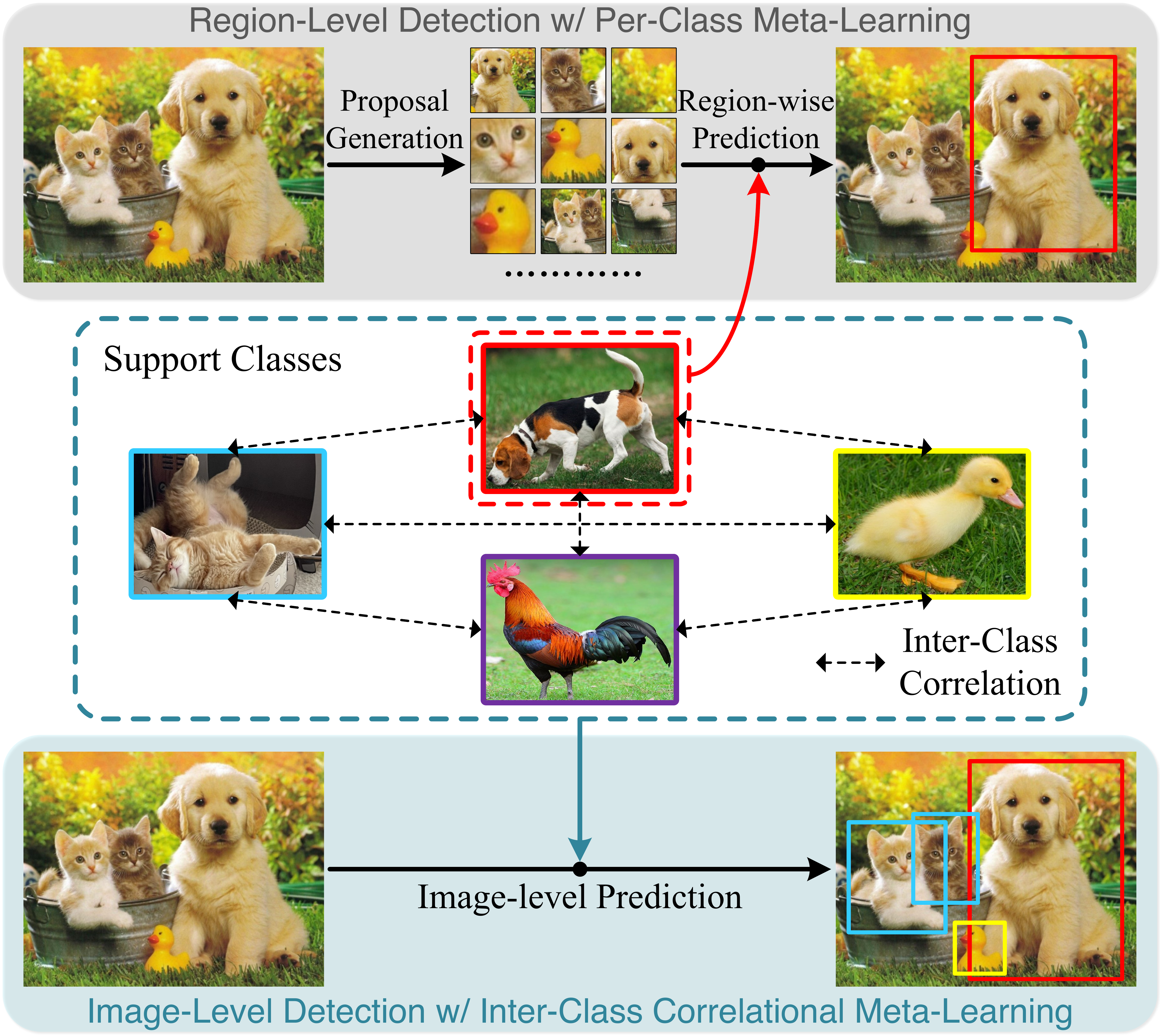}
\end{center}
\vspace{-3.88mm}
\caption{
Comparison of few-shot object detection pipelines: Prior studies (upper part) perform region-level detection, which are often constrained by inaccurate region proposals for novel classes. Besides, they can only deal with one support class at one go and overlook the correlation among different classes. The proposed Meta-DETR (lower part) works at image level without any proposals. It captures inter-class correlation by learning from multiple support classes simultaneously, which suppresses confusion among similar classes and enhances model generalization greatly.
}
\label{fig:fig1}
\vspace{-0.5mm}
\end{figure}

In this work, we explore the challenging task of \textit{few-shot object detection}, which requires detecting novel objects with only a few training samples. With minimal supervision on novel classes, the key to few-shot object detection is to learn transferable knowledge from base classes and generalize it to novel classes. To this end, many studies~\cite{FewshotReweighting,metarcnn,FSDetView,fsod,DenseRelationDistillation} incorporate meta-learning into generic region-based object detection frameworks, mostly Faster R-CNN~\cite{FasterRCNN}, and have achieved very promising results.

\begin{figure*}[t!] 
\begin{center}
   \includegraphics[width=1.0\linewidth]{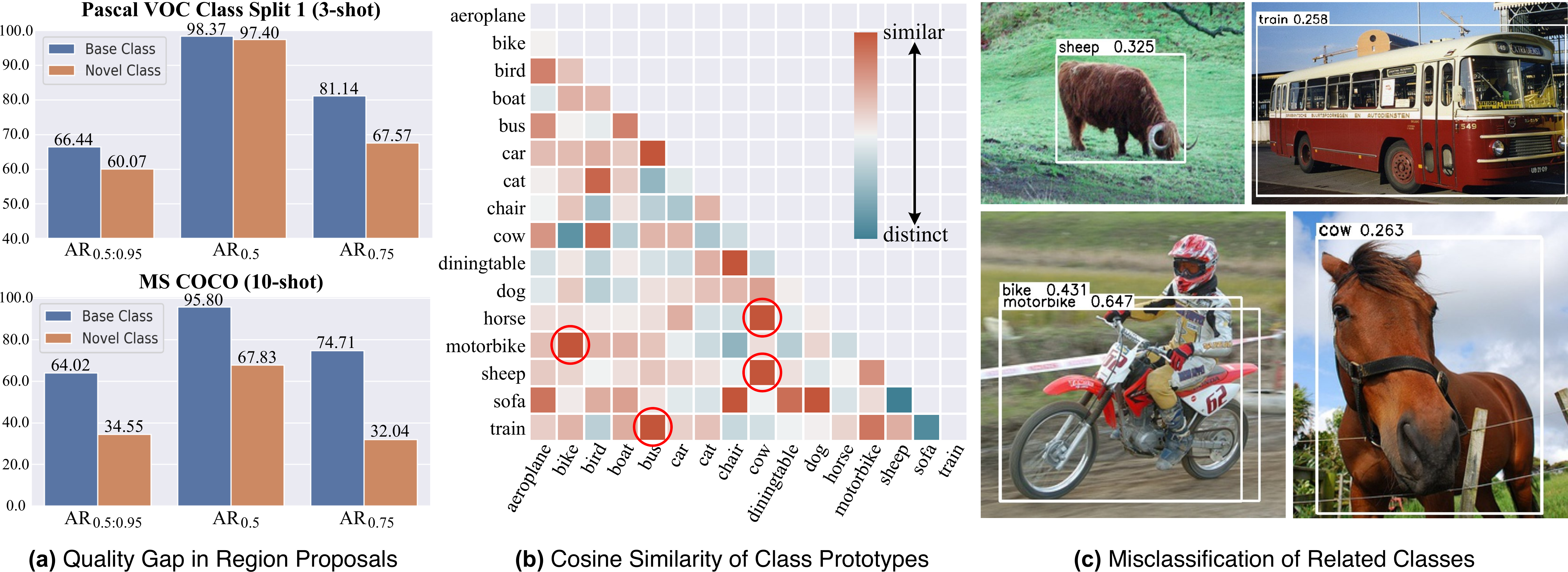}
\end{center}
\vspace{-4.486mm}
\caption{Existing few-shot detection frameworks tend to suffer from inaccurate region proposals and negligence of inter-class correlation. Due to very limited training samples for novel classes, the proposal quality (measured by Average Recall on top 1000 proposals) for novel classes is clearly lower than that of base classes as illustrated in (a). This hinders the knowledge generalization to novel classes. Additionally, object classes with similar appearances are highly correlated in feature space such as `cow \textit{vs.} horse' and `motorbike \textit{vs.} bike' as illustrated in (b), which tend to be misclassified if the learning does not incorporate the correlation among them as illustrated in (c).}
\label{fig:fig2_limitation}
\end{figure*}

Despite their success, there still exist two underlying limitations that hinder better exploitation of base-class knowledge, as illustrated in Fig.\,\ref{fig:fig2_limitation}.
\textit{First}, region-based detection frameworks rely on region proposals to produce final predictions, thus are sensitive to low-quality region proposals. However, as investigated by \cite{fsod} and \cite{CoRPN}, it is not easy to produce high-quality region proposals for novel classes with limited supervision under the few-shot detection setups. Such a gap in the quality of region proposals obstructs the generalization from base classes to novel classes.
\textit{Second}, most existing meta-learning-based approaches~\cite{FewshotReweighting,metarcnn,fsod,FSDetView} adopt `feature reweighting' or its variants to aggregate query and support features, which can only deal with one support class (\textit{i.e.}, target class to detect) at a time and essentially treat each support class independently.
Without seeing multiple classes within a single feedforward, they largely overlook the important inter-class correlation among different support classes. This limits the ability to distinguish similar classes (\textit{e.g.}, distinguishing from cows and horses) and to generalize from related classes (\textit{e.g.}, learning to detect cows by generalizing from detecting sheep).


To mitigate the above limitations, we design Meta-DETR, an innovative few-shot object detector that performs pure image-level prediction and at the same time exploits the inter-class correlation among different classes. Fig.~\ref{fig:fig1} illustrates its major differences with prior designs. To our best knowledge, this is the first work that identifies the constraint caused by region-based detection under the few-shot setups and explores to address few-shot object detection with DETR-based detection frameworks, which can skip proposal generation and directly perform detection at image level. With image-level prediction, Meta-DETR fully bypasses the constraint of inaccurate region proposals as in prevalent few-shot detection frameworks. In addition, the introduced inter-class correlational meta-learning strategy enables Meta-DETR to attend to multiple support classes at one go instead of class-by-class meta-learning with repeated runs as in most existing methods. By integrating detection tasks that involve multiple classes into meta-learning, Meta-DETR can explicitly leverage the inter-class correlation, including the inter-class commonality to facilitate generalization among related classes and the inter-class uniqueness to reduce misclassification among similar classes.

In summary, the contributions of this work are threefold.\\
\textit{First}, we identify the quality gap of proposals for base and novel classes in region-based prediction, and propose Meta-DETR to address few-shot object detection. Being the first pure image-level few-shot detector, Meta-DETR fully circumvents the gap of inaccurate proposals for novel-class objects, thus enabling better generalization to novel classes.
\textit{Second}, we design a novel correlational meta-learning strategy, which can deal with multiple support classes simultaneously. It effectively exploits inter-class correlation among different classes, thus greatly reducing misclassification and enhancing model generalization.
\textit{Third}, extensive experiments show that, without bells and whistles, the proposed Meta-DETR consistently outperforms state-of-the-art methods by large margins on detecting novel objects.

\begin{figure*}[t!] 
\begin{center}
   \includegraphics[width=1.0\linewidth]{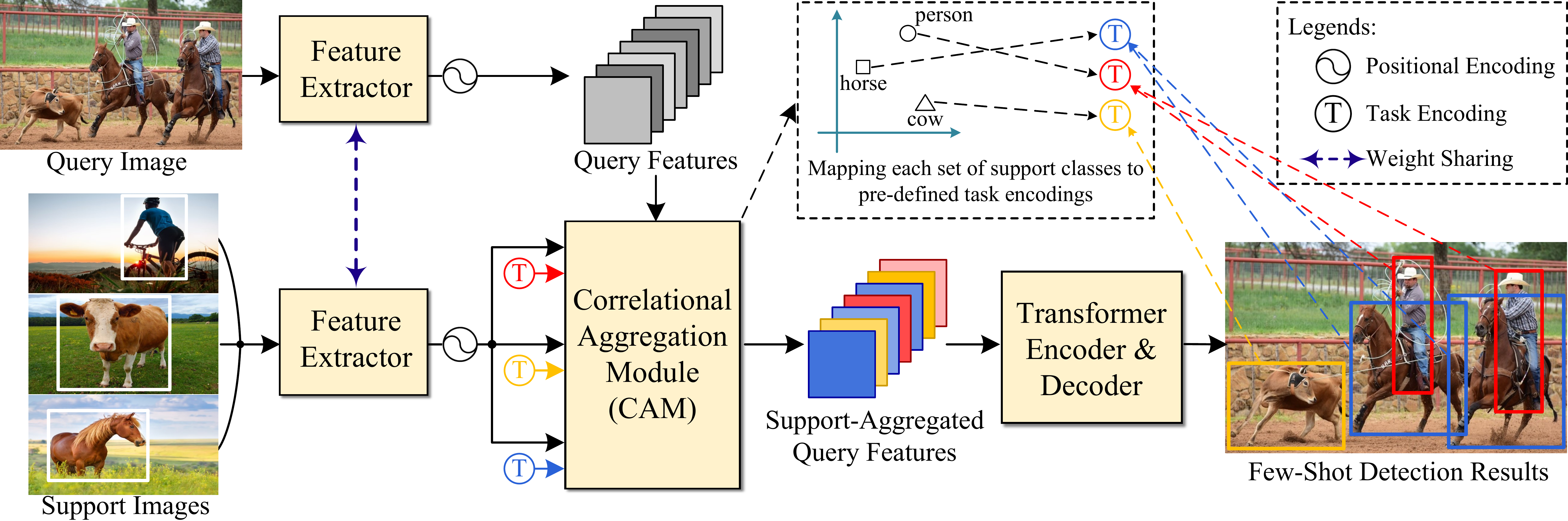}
\end{center}
\vspace{-4.486mm}
   \caption{
   The framework of Meta-DETR: \textit{Query Image} and \textit{Support Images} are processed by a weight-shared \textit{Feature Extractor} to produce \textit{Query Features} and support features. To leverage the inter-class correlation in meta-learning, a \textit{Correlational Aggregation Module (CAM)} is designed, which first matches the query features with multiple support classes simultaneously and then introduces multiple task encodings (\textit{i.e.}, the three illustrative \raisebox{-0.5pt}{\large \textcircled{\raisebox{-1.0pt} {\small T}}} of different colors) to differentiate these support classes. Finally, few-shot detection is achieved with a class-agnostic \textit{Transformer Encoder \& Decoder} that learns to predict objects' locations and their corresponding task encodings (instead of directly predicting objects' class labels). The architecture of CAM is detailed in Section\,\ref{sec:method_cam} and Fig.\,\ref{fig:fig4_CAM}. The training objectives of Meta-DETR are 
  formulated in Section\,\ref{sec:TrainingObjective}.
   }
\label{fig:fig3_architecture}
\end{figure*}

\section{Related Work}

\subsection{Object Detection}
Generic object detection~\cite{Liu2019DeepLF} is a joint task on object localization and classification. Modern object detectors are mostly region-based and can be broadly classified into two categories: two-stage and single-stage detectors. Two-stage detectors include Faster R-CNN~\cite{FasterRCNN} and its variants~\cite{relationnetworks_detection,CascadeRCNN,CADNet}, which first adopt a Region Proposal Network (RPN) to generate region proposals, and then produce final predictions based on the proposals. Differently, single-stage detectors~\cite{SSD,YOLO9000,RefineDet,zhou2019objects,incrementalfsdet,PNPDet} employ densely placed anchors as region proposals and directly make predictions over them.
Recently, another line of research featuring DETR~\cite{DETR} and its variants~\cite{DeformableDETR,Meta-DETR_firstversion,up-detr,SMCA-DETR,ConditionalDETR,DA-DETR,AnchorDETR,DABDETR,SAM-DETR,SAM-DETR++} has received vast attention, thanks to the merits of pure image-level framework, fully end-to-end pipeline, and comparable or even better performance.
However, these aforementioned generic detectors still heavily rely on large amounts of annotated training samples, thus will suffer from drastic performance drop when directly applied to few-shot object detection.

\subsection{Few-Shot Object Detection}
Existing works on few-shot object detection can be categorized into two paradigms: transfer learning and meta-learning. Methods with transfer learning mainly include LSTD~\cite{LSTD}, TFA~\cite{fsdet}, MPSR~\cite{MPSR}, and FSCE~\cite{fsce}, where novel concepts are learned via fine-tuning. Differently, methods with meta-learning~\cite{FewshotReweighting,metarcnn,metadet,incrementalfsdet,FSDetView,fsod,DenseRelationDistillation} extract knowledge that can generalize across various tasks via `learning to learn', \textit{i.e.}, learning a class-agnostic predictor on various auxiliary tasks.

Our proposed Meta-DETR falls under the umbrella of meta-learning, but differs from existing approaches by achieving image-level detection and effectively leveraging the correlation among various support classes. To the best of our knowledge, Meta-DETR is the first work that incorporates meta-learning into the recently proposed DETR frameworks. It is also the pioneering work to explicitly integrate the inter-class correlation among support classes into few-shot object detection frameworks using meta-learning.

\section{Preliminaries}

\subsection{Problem Definition}

Given two sets of classes $\mathcal{C}_{\rm base}$ and $\mathcal{C}_{\rm novel}$, where $\mathcal{C}_{\rm base} \cap \mathcal{C}_{\rm novel}=\varnothing$, a few-shot object detector aims at detecting objects of $\mathcal{C}_{\rm base} \cup \mathcal{C}_{\rm novel}$ by learning from a base dataset $\mathcal{D}_{\rm base}$ with abundant annotated objects of $\mathcal{C}_{\rm base}$ and a novel dataset $\mathcal{D}_{\rm novel}$ with very few annotated objects of $\mathcal{C}_{\rm novel}$. In the task of $K$-shot object detection, there are exactly $K$ annotated objects for each novel class in $\mathcal{D}_{\rm novel}$.

\subsection{Rethink Region-Based Detection Frameworks}

Most existing few-shot object detectors are developed on top of Faster R-CNN~\cite{FasterRCNN}, a region-based object detector, thanks to its robust performance and ease for optimization. However, by relying on region proposals to produce detection results, these approaches are inevitably constrained by the inaccurate proposals for novel classes due to very limited supervision under the few-shot detection setups. As illustrated in Fig.\;\ref{fig:fig2_limitation}(a), there is a clear gap in the quality of region proposals for base and novel classes, hindering region-based detection frameworks from exploiting base-class knowledge to generalize to novel classes. Though several studies~\cite{fsod,CoRPN} attempt to acquire more accurate region proposals, this issue still remains as it is rooted in the region-based detection frameworks under the few-shot learning setups.

\subsection{Rethink Meta-Learning via Feature Reweighting}

To meta-learn a class-agnostic detector that can generalize across various classes, most existing methods~\cite{FewshotReweighting,metarcnn,fsod,FSDetView} adopt `feature reweighting' or its variants to aggregate query features with support class information, acquiring class-specific meta-features to detect objects corresponding to the support class. However, such meta-learning strategies can deal with only one support class within each feed-forward process, \textit{i.e.}, $C$ repeated runs are required to detect $C$ support classes within each query image. More importantly, by treating each support class independently, `feature reweighting' overlooks the essential inter-class correlation among different support classes. As shown in Fig.\,\ref{fig:fig2_limitation}(b), many object classes with similar appearances are highly correlated. Intuitively, their correlation can effectively facilitate the distinction and the generalization among similar classes. However, as shown in Fig.\,\ref{fig:fig2_limitation}(c), we observe that objects misclassified as highly correlated classes constitute a major source of error due to the negligence of inter-class correlation in existing methods.

\vspace{-1.5mm}
\section{Meta-DETR}  \label{sec:method}

This section provides a detailed description of the proposed Meta-DETR, including its network architecture, training objective, as well as the learning and inference procedure.

\vspace{-1.5mm}
\subsection{Model Overview}   \label{sec:method_overview}

Fig.\,\ref{fig:fig3_architecture} shows the architecture of the proposed Meta-DETR. Motivated by previous discussions, Meta-DETR employs the recently proposed Deformable DETR~\cite{DeformableDETR}, a fully end-to-end Transformer-based~\cite{transformer} detector, as the basic detection framework. As Meta-DETR does not rely on predicted region proposals to make final predictions, it can fully bypass the constraint of inaccurate proposals on novel-class objects. Besides, thanks to the introduced correlational meta-learning, Meta-DETR can aggregate query features with multiple support classes simultaneously, thus capturing and leveraging the inter-class correlation among different classes to reduce misclassification and boost generalization.


Given a query image and a set of support images with instance annotations, a weight-shared feature extractor first encodes them into the same feature space. Subsequently, a \textit{Correlational Aggregation Module (CAM)}, which will be introduced later, performs simultaneous aggregation between the query features and the set of support classes. To differentiate between different support classes in a class-agnostic manner, CAM introduces a set of task encodings assigned to each support class. Finally, a transformer architecture detects objects by predicting their locations and corresponding task encodings. As the detection targets are dynamically determined by support classes and their mappings to task encodings, Meta-DETR is trained as a meta-learner to extract generalizable knowledge not specific to certain classes.

\subsection{Inter-Class Correlational Meta-Learning}   \label{sec:method_cam}

\begin{figure}[t!] 
\begin{center}
   \includegraphics[width=1.0\linewidth]{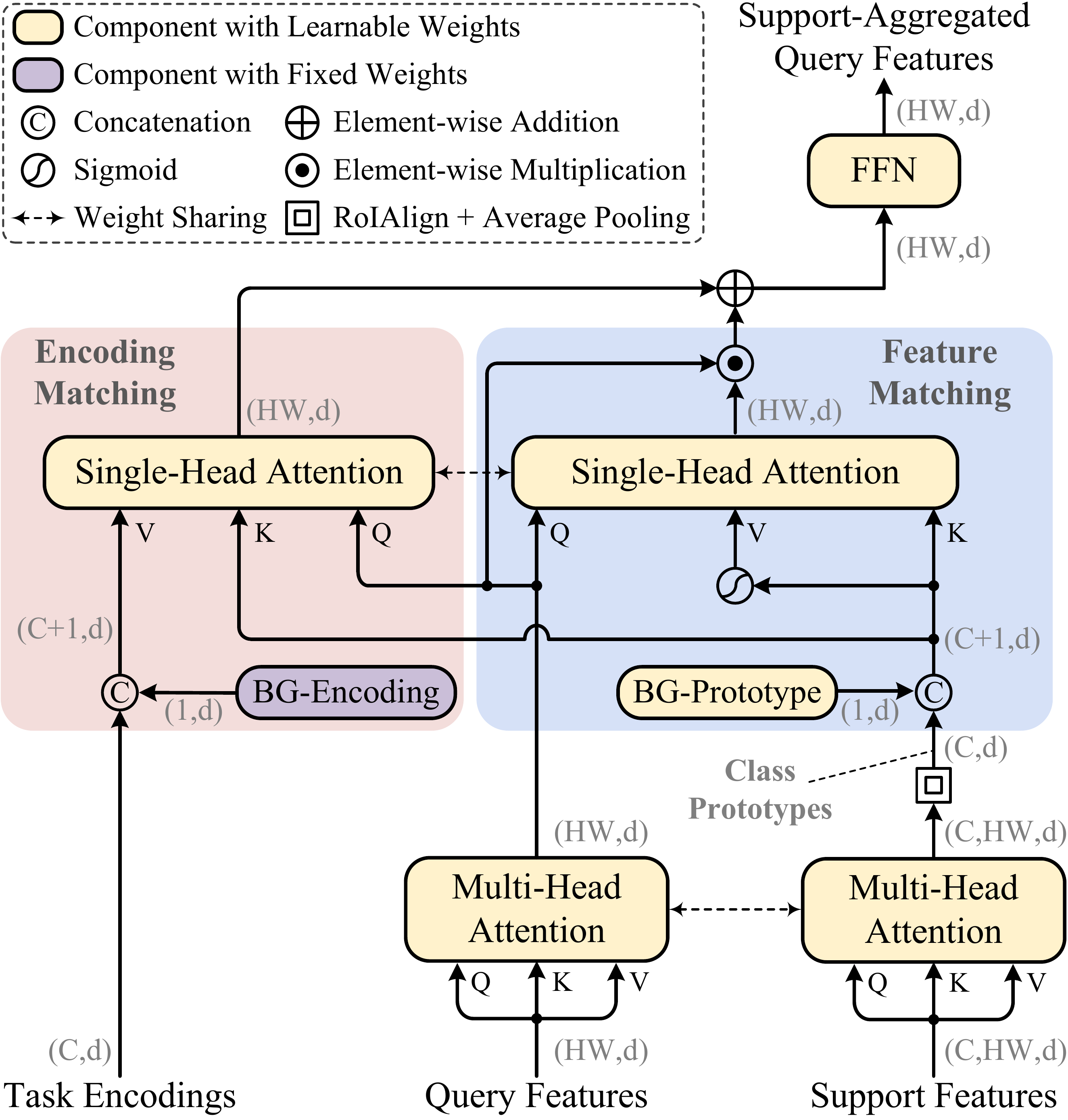}
\end{center}
\vspace{-4.486mm}
   \caption{
   The architecture of the Correlational Aggregation Module (CAM). CAM first obtains class prototypes from support features. Then, it performs two matching processes: \textit{Feature Matching} filters out query features that are unrelated to support classes, while \textit{Encoding Matching} matches query features to a set of pre-defined task encodings that differentiate their corresponding support classes in a class-agnostic manner.
   }
\label{fig:fig4_CAM}
\end{figure}

The \textit{Correlational Aggregation Module (CAM)} is the key component in Meta-DETR to perform inter-class correlational meta-learning, which aggregates query features with support classes for the subsequent class-agnostic prediction.
CAM differs from existing aggregation methods in that it can aggregate multiple support classes simultaneously, which enables it to capture their inter-class correlation to reduce misclassification and enhance model generalization. Specifically, as illustrated in Fig.\;\ref{fig:fig4_CAM}, the query and support features are first processed by a weight-shared multi-head attention module, encoding them into the same embedding space. Then the prototype for each support class is obtained by applying RoIAlign~\cite{MaskRCNN}, followed by average pooling on the support features, where RoIAlign ensures that class prototypes are obtained from the relevant regions that contain corresponding support object instances. After that, CAM performs feature matching and encoding matching, which will be elaborated in the remainder of this subsection to match the query features with support class prototypes and task encodings, respectively. The matching results are summed together and fed to a feed-forward network (FFN) to produce the final output. Note that the support class prototypes are obtained in CAM before feature matching and encoding matching.


\subsubsection{Feature Matching}

Feature matching, which aims to filter out features irrelevant to support classes, is achieved by an attention mechanism with minor modifications. Specifically, given a query feature map $\mathbf{Q} \in \mathbb{R}^{HW \times d}$ and the support class prototypes $\mathbf{S} \in \mathbb{R}^{C \times d}$, where $HW$ is the spatial size, $C$ is the number of support classes, and $d$ is the feature dimensionality, the matching coefficients are obtained via:
\begin{equation}
\mathbf{A} = {\rm Attn}(\mathbf{Q}, \mathbf{S}) = {\rm Softmax}(\frac{(\mathbf{Q W})(\mathbf{S W})^{\rm T}}{\sqrt{d}}),
\end{equation}
where $\mathbf{W}$ is a linear projection shared by $\mathbf{Q}$ and $\mathbf{S}$, which ensures they are embedded into the same feature space. Subsequently, the output of the feature matching module can be obtained via:
\begin{equation}
\mathbf{Q}_{\mathbf{F}} = \mathbf{A} \sigma(\mathbf{S}) \odot \mathbf{Q},
\end{equation}
where $\sigma(\cdot)$ denotes sigmoid function and $\odot$ denotes Hadamard product. $\sigma(\mathbf{S})$ serves as feature filters for each individual support class with the function of extracting only class-related features from query features. By applying the matching coefficients $\mathbf{A}$ to $\sigma(\mathbf{S})$, we have filters that can filter out query features that are not matched to any support class, producing a filtered query feature map $\mathbf{Q_F}$ that only highlights objects belonging to the given support classes.

\subsubsection{Encoding Matching}

To achieve correlational meta-learning, we introduce a set of pre-defined task encodings assigned to each support class and match query features to their corresponding task encodings, so that final predictions can be made on the task encodings instead of specific classes. We implement task encodings $\mathbf{T} \in \mathbb{R}^{C \times d}$ with sinusoidal functions, following the positional encodings of the Transformer~\cite{transformer}. Encoding matching uses the same matching coefficients as feature matching, and the matched encodings $\mathbf{Q_E}$ are obtained via:
\begin{equation}
\mathbf{Q}_{\mathbf{E}} = \mathbf{A} \mathbf{T}.
\end{equation}

\subsubsection{Modeling Background for Open-Set Prediction}

Object detection features an open-set setup where background, which does not belong to any of the target classes, often takes up most of the spatial locations in a query image. Therefore, as shown in Fig.\,\ref{fig:fig4_CAM}, we additionally introduce a learnable prototype and a corresponding task encoding (fixed to zeros), denoted as BG-Prototype and BG-Encoding respectively, to explicitly model the background class. This eliminates the matching ambiguity when query does not match any of the given support classes.

\subsection{Training Objective} \label{sec:TrainingObjective}

\subsubsection{Target Generation}
We let $N$ denote the fixed number of object queries, which means Meta-DETR infers $N$ predictions within a single feed-forward process. Let $x_{\rm query}$ denote the query image, and $y \! = \! \left\{ y_i \right\}_{i=1}^{N}$ denote the ground truth objects within the query image, where $y$ is a set of size $N$. When $y_i$ indicates an object, $y_i \! = \! (c_i, b_i)$, where $c_i$ denotes the target class label and $b_i$ denotes the bounding box of the object. When $y_i$ indicates no object, $y_i \! = \! (\varnothing, \varnothing)$.

Meta-DETR dynamically conditions its detection targets on the sampled support classes and their mappings to the task encodings. As discussed in Section~\ref{sec:method_overview}, Meta-DETR predicts over $C$ support classes (\textit{i.e.}, target classes) simultaneously. The $C$ support classes are randomly sampled, denoted as $c_{\rm supp} \! = \! \left\{ s_i \right\}_{i=1}^{C}$. Besides, these support classes are further mapped to a set of task encodings. We denote the mapping function from the labels of support classes to the labels of task encodings as $\chi(\cdot)$. A specific case of $\chi(\cdot)$ can be formulated as:
\begin{equation}
\chi(s_i) = i \quad\quad i \in \{1,2, \cdots, C\}.
\end{equation}
Note that the exact format of the mapping function $\chi(\cdot)$ does not matter.
Then, the detection targets of Meta-DETR can be formulated as:
\begin{equation}
y^{\prime} = \left\{ y^{\prime}_i \right\}_{i=1}^{N} = \left\{ (c^{\prime}_i, b^{\prime}_i) \right\}_{i=1}^{N} = \left\{ \psi(y_i, c_{\rm supp}) \right\}_{i=1}^{N},
\end{equation}
where $\psi(y_i, c_{\rm supp})$ acts to remove annotations of irrelevant objects (objects with labels not in $c_{\rm supp}$) and to map the labels of target classes to the labels of the corresponding task encodings, which can be formulated as:
\begin{equation}
	\psi(y_i, c_{\rm supp})=\left\{
		\begin{aligned}
		&(\varnothing, \varnothing),& {\rm if}& \; y_i = (\varnothing, \varnothing) \;\, \rm{or} \;\, c_i \notin c_{\rm supp} &\\ 
		&(\chi(c_{i}), b_i),& {\rm if}& \; c_i \in c_{\rm supp}.& \\ 
	\end{aligned}
	\right.
\end{equation}
Note that $y^{\prime}$ can completely consist of $(\varnothing, \varnothing)$ when there is no objects that belong to the provided support classes.

\subsubsection{Loss Function}
Assume the $N$ predictions for target class made by Meta-DETR are $\hat{y} \! = \! \left\{ \hat{y}_i \right\}_{i=1}^{N} \! = \! \big\{ (\hat{c}_{i}, \hat{b}_{i}) \big\}_{i=1}^{N}$. We adopt a pair-wise matching loss $\mathcal{L}_{\rm match}(y^{\prime}_i, \hat{y}_{\sigma(i)})$ to search for a bipartite matching between $\hat{y}$ and $y^{\prime}$ with the lowest cost:
\begin{equation} \label{eq:optimalassignment}
\hat{\sigma} = \mathop{\arg\min}_{\sigma} \sum\nolimits_{i=1}^{N}\mathcal{L}_{\rm match}(y^{\prime}_i, \hat{y}_{\sigma(i)}),
\end{equation}
where $\sigma$ denotes a permutation of $N$ elements, and $\hat{\sigma}$ denotes the optimal assignment between predictions and targets. Since the matching should consider both classification and localization, the matching loss is defined as:
\begin{equation} \label{eq:Lmatch}
\begin{split}
\mathcal{L}_{\rm match}(y^{\prime}_i, \hat{y}_{\sigma(i)}) = &\mathbbm{1}_{\left\{ c^{\prime}_i \neq \varnothing \right\}} \mathcal{L}_{\rm cls}(c^{\prime}_i, \hat{c}_{\sigma(i)}) \, + \\&\mathbbm{1}_{\left\{ c^{\prime}_i \neq \varnothing \right\}} \mathcal{L}_{\rm box}(b^{\prime}_i, \hat{b}_{\sigma(i)}) \;\;.
\end{split}
\end{equation}

With the optimal assignment $\hat{\sigma}$ obtained with Eq.\,\ref{eq:optimalassignment} and Eq.\,\ref{eq:Lmatch}, we optimize the network using the following loss function:
\begin{equation} \label{eq:L}
\begin{split}
\mathcal{L}(y^{\prime}, \hat{y}) \! = \! \sum\limits_{i=1}^{N} \left[ \mathcal{L}_{\rm cls}(c^{\prime}_i, \hat{c}_{\hat{\sigma}(i)}) + \mathbbm{1}_{\left\{ c^{\prime}_i \neq \varnothing \right\}} \mathcal{L}_{\rm box}(b^{\prime}_i, \hat{b}_{\hat{\sigma}(i)}) \right],
\end{split}
\end{equation}
where we adopt sigmoid focal loss~\cite{focalloss} for $\mathcal{L}_{\rm cls}$ and adopt a linear combination of $\ell1$ loss and GIoU loss~\cite{giouloss} for $\mathcal{L}_{\rm box}$. Similar to DETR~\cite{DETR} and Deformable DETR~\cite{DeformableDETR}, $\mathcal{L}(y^{\prime}, \hat{y})$ is applied to every layer of the transformer decoder.

Following Meta R-CNN~\cite{metarcnn}, we introduce a cosine similarity cross-entropy loss~\cite{CloserFewshotClassification} to classify the class prototypes obtained by our designed CAM. It encourages prototypes of different classes to be distinguished from each other.

\begin{table*}[t]
\begin{center}
\centering
\caption{Few-shot detection performance (mAP@0.5) on Pascal VOC for novel classes}
\vspace{-4.50mm}
\label{tab:Performance_VOC_novel}
\setlength{\tabcolsep}{4.75pt}
\resizebox{1.0\textwidth}{!}{
\begin{tabular}[t]{l|ccccc|ccccc|ccccc|c}
\toprule[1.1pt]
&\multicolumn{5}{c|}{Class Split 1} & \multicolumn{5}{c|}{Class Split 2} & \multicolumn{5}{c|}{Class Split 3} & \multirow{2}*{\textit{Avg.}} \\

 \cmidrule{1-16}

Method $\backslash$ Shots  & 1 & 2 & 3 & 5 & 10 & 1 & 2 & 3 & 5 & 10 & 1 & 2 & 3 & 5 & 10 & \\\midrule[0.88pt]

\multicolumn{17}{l}{\textit{Results over a single run:}} \\\midrule[0.486pt]

LSTD~\cite{LSTD} & 8.2 & 1.0 & 12.4 & 29.1 & 38.5 & 11.4 & 3.8 & 5.0 & 15.7 & 31.0 & 12.6 & 8.5 & 15.0 & 27.3 & 36.3 & 17.1 \\

RepMet~\cite{RepMet}\,$\ddag$ & 26.1 & 32.9 & 34.4 & 38.6 & 41.3 & 17.2 & 22.1 & 23.4 & 28.3 & 35.8 & 27.5 & 31.1 & 31.5 & 34.4 & 37.2 & 30.8 \\


Meta-YOLO~\cite{FewshotReweighting} & 14.8 & 15.5 & 26.7 & 33.9 & 47.2 & 15.7 & 15.3 & 22.7 & 30.1 & 40.5 & 21.3 & 25.6 & 28.4 & 42.8 & 45.9 & 28.4 \\

Meta\,Det~\cite{metadet} & 18.9 & 20.6 & 30.2 & 36.8 & 49.6 & 21.8 & 23.1 & 27.8 & 31.7 & 43.0 & 20.6 & 23.9 & 29.4 & 43.9 & 44.1 & 31.0 \\

Meta R-CNN~\cite{metarcnn} & 19.9 & 25.5 & 35.0 & 45.7 & 51.5 & 10.4 & 19.4 & 29.6 & 34.8 & 45.4 & 14.3 & 18.2 & 27.5 & 41.2 & 48.1 & 31.1 \\

TFA w/ fc~\cite{fsdet}\,$\ddag$ & 36.8 & 29.1 & 43.6 & 55.7 & 57.0 & 18.2 & 29.0 & 33.4 & 35.5 & 39.0 & 27.7 & 33.6 & 42.5 & 48.7 & 50.2 & 38.7 \\

TFA w/ cos~\cite{fsdet}\,$\ddag$ & 39.8 & 36.1 & 44.7 & 55.7 & 56.0 & 23.5 & 26.9 & 34.1 & 35.1 & 39.1 & 30.8 & 34.8 & 42.8 & 49.5 & 49.8 & 39.9 \\

MPSR~\cite{MPSR}\,$\ddag$ & 41.7 & 43.1 & 51.4 & 55.2 & 61.8 & 24.4 & 29.5 & 39.2 & 39.9 & 47.8 & 35.6 & 40.6 & 42.3 & 48.0 & 49.7 & 43.3 \\

TFA w/ cos + Halluc~\cite{Halluc_FSD}\,$\ddag$ & 45.1 & 44.0 & 44.7 & 55.0 & 55.9 & 23.2 & 27.5 & 35.1 & 34.9 & 39.0 & 30.5 & 35.1 & 41.4 & 49.0 & 49.3 & 40.6 \\

Retentive R-CNN~\cite{retentive_rcnn}\,$\ddag$ & 42.4 & 45.8 & 45.9 & 53.7 & 56.1 & 21.7 & 27.8 & 35.2 & 37.0 & 40.3 & 30.2 & 37.6 & 43.0 & 49.7 & 50.1 & 41.1 \\

CME~\cite{CME}\,$\ddag$ & 41.5 & 47.5 & 50.4 & 58.2 & 60.9 & 27.2 & 30.2 & 41.4 & 42.5 & 46.8 & 34.3 & 39.6 & 45.1 & 48.3 & 51.5 & 44.4 \\

SRR-FSD~\cite{SRR-FSD}\,$\ddag$\,$\uplus$ & \textbf{47.8} & 50.5 & 51.3 & 55.2 & 56.8 & 32.5 & 35.3 & 39.1 & 40.8 & 43.8 & 40.1 & 41.5 & 44.3 & 46.9 & 46.4 & 44.8\\

FSCE~\cite{fsce}\,$\ddag$ & 44.2 & 43.8 & 51.4 & \textbf{61.9} & 63.4 & 27.3 & 29.5 & 43.5 & 44.2 & 50.2 & 37.2 & 41.9 & 47.5 & 54.6 & 58.5 & 46.6\\

\rowcolor{black!6} Meta-DETR (Ours) & 40.6 & \textbf{51.4} & \textbf{58.0} & 59.2 & \textbf{63.6} & \textbf{37.0} & \textbf{36.6} & \textbf{43.7} & \textbf{49.1} & \textbf{54.6} & \textbf{41.6} & \textbf{45.9} & \textbf{52.7} & \textbf{58.9} & \textbf{60.6} & \textbf{50.2} \\\midrule[0.88pt]

\multicolumn{17}{l}{\textit{Results averaged over multiple random runs:}} \\\midrule[0.486pt]

FRCN+ft-full~\cite{FasterRCNN}\,$\ddag$ & 9.9 & 15.6 & 21.6 & 28.0 & 35.6 & 9.4 & 13.8 & 17.4 & 21.9 & 29.8 & 8.1 & 13.9 & 19.0 & 23.9 & 31.0 & 19.9 \\

Deformable-DETR+ft-full~\cite{DeformableDETR}\,$\ddag$ & 5.6 & 13.3 & 21.7 & 34.2 & 45.0 & 10.9 & 13.0 & 18.4 & 27.3 & 39.4 & 7.3 & 16.6 & 20.8 & 32.2 & 41.8 & 23.2 \\

TFA w/ fc~\cite{fsdet}\,$\ddag$ & 22.9 & 34.5 & 40.4 & 46.7 & 52.0 & 16.9 & 26.4 & 30.5 & 34.6 & 39.7 & 15.7 & 27.2 & 34.7 & 40.8 & 44.6 & 33.8 \\

TFA w/ cos~\cite{fsdet}\,$\ddag$ & 25.3 & 36.4 & 42.1 & 47.9 & 52.8 & 18.3 & 27.5 & 30.9 & 34.1 & 39.5 & 17.9 & 27.2 & 34.3 & 40.8 & 45.6 & 34.7 \\

FsDetView~\cite{FSDetView} & 24.2 & 35.3 & 42.2 & 49.1 & 57.4 & 21.6 & 24.6 & 31.9 & 37.0 & 45.7 & 21.2 & 30.0 & 37.2 & 43.8 & 49.6 & 36.7 \\

MPSR~\cite{MPSR}\,$\ddag\,{\triangle}$ & 34.7 & 42.6 & 46.1 & 49.4 & 56.7 & 22.6 & 30.5 & 31.0 & 36.7 & 43.3 & 27.5 & 32.5 & 38.2 & 44.6 & 50.0 & 39.1 \\

DCNet~\cite{DenseRelationDistillation}\,$\ddag$ & 33.9 & 37.4 & 43.7 & 51.1 & 59.6 & 23.2 & 24.8 & 30.6 & 36.7 & 46.6 & 32.3 & 34.9 & 39.7 & 42.6 & 50.7 & 39.2 \\

FSCE~\cite{fsce}\,$\ddag$ & 32.9 & 44.0 & 46.8 & 52.9 & 59.7 & 23.7 & 30.6 & \textbf{38.4} & 43.0 & 48.5 & 22.6 & 33.4 & 39.5 & 47.3 & 54.0 & 41.2 \\

\rowcolor{black!6} Meta-DETR (Ours) & \textbf{35.1} & \textbf{49.0} & \textbf{53.2} & \textbf{57.4} & \textbf{62.0} & \textbf{27.9} & \textbf{32.3} & \textbf{38.4} & \textbf{43.2} & \textbf{51.8} & \textbf{34.9} & \textbf{41.8} & \textbf{47.1} & \textbf{54.1} & \textbf{58.2} & \textbf{45.8} \\

\bottomrule[1.1pt]

\end{tabular}}
\vspace{+0.25mm}
  \begin{tablenotes}
    \item[1] "\textit{$\ddag$" indicates methods using multi-scale features. \quad\quad\quad "$\triangle$" indicates re-evaluated results using official codes. \quad\quad\quad "$\uplus$" indicates usage of external data.}
  \end{tablenotes}
\vspace{-2.30mm}
\end{center}
\end{table*}

\begin{table}[t]
\begin{center}
\centering
\caption{Few-shot detection performance (mAP@0.5) \\ on Pascal VOC class split 1 for both base and novel classes}
\vspace{-4.0mm}
\label{tab:Performance_VOC1_basenovel}
\setlength{\tabcolsep}{3.25pt}
\resizebox{0.49\textwidth}{!}{
\begin{tabular}[t]{l|cccc|cccc}
\toprule[1.0pt]

&\multicolumn{4}{c|}{Base Classes} & \multicolumn{4}{c}{Novel Classes}\\

Method $\backslash$ Shots & 1 & 3 & 5 & 10 & 1 & 3 & 5 & 10 \\

\midrule[0.68pt]

Meta-YOLO~\cite{FewshotReweighting} & 66.4 & 64.8 & 63.4 & 63.6 & 14.8 & 26.7 & 33.9 & 47.2 \\

FsDetView~\cite{FSDetView}\,$\S$ & 64.2 & 69.4 & 69.8 & 71.1 & 24.2 & 42.2 & 49.1 & 57.4 \\

TFA\,w/\,cos~\cite{fsdet}\,$\S$ & \textbf{77.6} & \textbf{77.3} & \textbf{77.4} & \textbf{77.5} & 25.3 & 42.1 & 47.9 & 52.9\\

MPSR~\cite{MPSR}\,$\S$ & 60.6 & 65.9 & 68.2 & 69.8 & 34.7 & 46.1 & 49.4 & 56.7 \\

FSCE~\cite{fsce}\,$\S$ & 75.5 & 73.7 & 75.0 & 75.2 & 32.9 & 46.8 & 52.9 & 59.7 \\

\rowcolor{black!6} Meta-DETR\,(Ours)\,$\S$ & 67.2 & 70.0 & 73.0 & 73.5 & \textbf{35.1} & \textbf{53.2} & \textbf{57.4} & \textbf{62.0} \\

\bottomrule[1.0pt]

\end{tabular}
}
\vspace{+0.25mm}
  \begin{tablenotes}
    \item[1] \textit{"$\S$" indicates results averaged over multiple random runs.}
  \end{tablenotes}
  
\vspace{-2.50mm}  
\end{center}
\end{table}

\subsection{Training and Inference Procedure}

\subsubsection{Two-Stage Training Procedure}
The training procedure consists of two stages.
The first stage is \textit{base training stage}. During this stage, the model is trained on the base dataset $\mathcal{D}_{\rm base}$ with abundant training samples for each base class.
The second stage is \textit{few-shot fine-tuning stage}. In this stage, we train the model on both base and novel classes with limited training samples. Only $K$ object instances are available for each novel category in $K$-shot object detection. Following prior works~\cite{metarcnn,fsdet,FSDetView}, we also include objects from base classes to prevent performance drop for base classes. In both \textit{base training} and \textit{few-shot fine-tuning} stages, the whole network is optimized in an end-to-end manner with the same training objective described in Section\;\ref{sec:TrainingObjective}.

\subsubsection{Efficient Inference}
Unlike the training stage, there is no need to repeatedly sample support images and extract their features with the feature extractor. We can first compute the prototype for each support class once and for all, then directly use them for every query image to predict. This promises efficient inference of our proposed Meta-DETR.

\section{Experiments}

\subsection{Datasets}  \label{sec:datasets}

We follow the well-established data setups for few-shot object detection~\cite{FewshotReweighting,metadet,metarcnn,fsdet,FSDetView}. Concretely, two widely used few-shot object detection benchmarks are adopted in our experiments.

\smallskip
\textbf{Pascal VOC}~\cite{PascalVOC}
is a commonly used dataset for object detection that consists of images with object annotations of 20 classes. We use \textit{trainval\,07+12} for training and perform evaluations on \textit{test\,07}. We use 3 novel\,/\,base class splits, \textit{i.e.}, (“bird”, “bus”, “cow”, “motorbike”, “sofa”\;/\;others), (“aeroplane”, “bottle”,“cow”,“horse”,“sofa”\;/\;others), and (“boat”, “cat”, “motorbike”,“sheep”, “sofa”\;/\;others). The number of shots is set to 1, 2, 3, 5 and 10.  Mean average precision at IoU threshold 0.5 (mAP@0.5) is used as the evaluation metric. Results are averaged over 10 randomly sampled support datasets.

\smallskip
\textbf{MS\;COCO}\,\cite{MSCOCO}
is a more challenging object detection dataset, which contains 80 classes including those 20 classes in Pascal\;VOC. We adopt the 20 shared classes as novel classes, and adopt the remaining 60 classes as base classes. The number of shots is set to 1, 3, 5, 10, and 30. We use \textit{train\,2017} for training, and perform evaluations on \textit{val\,2017}. Standard evaluation metrics for MS\;COCO are adopted. Results are averaged over 5 randomly sampled support datasets.

\vspace{-2.0mm}
\subsection{Implementation Details} \label{sec:imple_details}

We adopt the commonly used ResNet-101~\cite{resnet} as the feature extractor. The network architectures and hyper-parameters remain the same as Deformable DETR\;\cite{DeformableDETR}. We implement our model in single-scale version for fair comparison with other works. We also follow FsDetView~\cite{FSDetView} to implement the aggregation with a slightly more complex scheme compared with solely feature reweighting. We train our model with 8\,x Nvidia V100 GPUs, using the AdamW~\cite{AdamW} optimizer with an initial learning rate of $\rm 2\!\times\!10^{-4}$ and a weight decay of $\rm 1\!\times\!10^{-4}$. Batch size is set to 32. In the base training stage, we train the model for 50 and 25 epochs for Pascal VOC and MS COCO, respectively. Learning rate is decayed at the 45$^{\rm th}$ and 20$^{\rm th}$ epoch by 0.1. In the few-shot fine-tuning stage, the same settings are applied to fine-tune the model until convergence.

\vspace{-2.0mm}
\subsection{Comparison with State-of-the-Art Methods}

\subsubsection{Pascal VOC}

Table\;\ref{tab:Performance_VOC_novel} shows the few-shot detection performance for novel classes of Pascal VOC. It can be seen that Meta-DETR consistently outperforms existing methods across various setups. With multiple runs over randomly sampled support datasets to reduce randomness, Meta-DETR achieves the best average performance across all setups, with a large margin of +\,4.6\% overall mAP compared with the second-best. The strong performance demonstrates the superiority and robustness of our proposed Meta-DETR.

We also present results taking base classes into consideration in Table\;\ref{tab:Performance_VOC1_basenovel}. While achieving good performance for novel classes with limited training samples, Meta-DETR can still detect objects of base classes with competitive performance. TFA~\cite{fsdet} produces outstanding performance for base classes since it only fine-tunes detector's last layer, thus having relatively constrained capacity in generalizing on novel classes. We would highlight that our proposed Meta-DETR achieves the best base-class and novel-class performance among all compared methods using meta-learning (\textit{i.e.}, Meta-YOLO~\cite{FewshotReweighting} and FsDetView~\cite{FSDetView}).

\begin{table}[t!]
\begin{center}
\centering
\caption{Few-shot detection performance on COCO for novel classes}
\label{tab:Performance_COCO_novel}
\vspace{-3.8mm}
\setlength{\tabcolsep}{2.486pt}
\resizebox{0.486\textwidth}{!}{
\begin{tabular}[t]{ c | l | ccc }
\toprule[1.0pt]

Shot & Method & AP$_{\rm 0.5:0.95}$ & AP$_{\rm 0.5}$ & AP$_{\rm 0.75}$  \\

\midrule[0.68pt]

\multirow{5}{*}{1} & FRCN+ft-full~\cite{FasterRCNN}\,$\ddag$\,$\S$ & 1.7 & 3.3 & 1.6 \\

& Deformable-DETR+ft-full~\cite{DeformableDETR}\,$\S$ & 1.8 & 3.1 & 1.8 \\

& TFA w/ cos~\cite{fsdet}\,$\ddag$\,$\S$ & 1.9 & 3.8 & 1.7 \\

& TFA\,w/\,cos\,+\,Halluc~\cite{Halluc_FSD}\,$\ddag$ & 3.8 & 6.5 & 4.3 \\

& \cellcolor{black!6}Meta-DETR (Ours)\,$\S$ & \cellcolor{black!6}\textbf{7.5} & \cellcolor{black!6}\textbf{12.5} & \cellcolor{black!6}\textbf{7.7} \\

\midrule[0.68pt]

\multirow{5}{*}{3} & FRCN+ft-full~\cite{FasterRCNN}\,$\ddag$\,$\S$ & 3.7 & 7.1 & 3.5 \\

& Deformable-DETR+ft-full~\cite{DeformableDETR}\,$\S$ & 4.9 & 7.8 & 5.1 \\

& TFA w/ cos~\cite{fsdet}\,$\ddag$\,$\S$ & 5.1 & 9.9 & 4.8 \\

& TFA\,w/\,cos\,+\,Halluc~\cite{Halluc_FSD}\,$\ddag$ & 6.9 & 12.6 & 7.0 \\

& \cellcolor{black!6}Meta-DETR (Ours)\,$\S$ & \cellcolor{black!6}\textbf{13.5} & \cellcolor{black!6}\textbf{21.7} & \cellcolor{black!6}\textbf{14.0} \\

\midrule[0.68pt]

\multirow{5}{*}{5} & FRCN+ft-full~\cite{FasterRCNN}\,$\ddag$\,$\S$ & 4.6 & 8.7 & 4.4 \\

& Deformable-DETR+ft-full~\cite{DeformableDETR}\,$\S$ & 7.4 & 12.3 & 7.7 \\

& TFA w/ cos~\cite{fsdet}\,$\ddag$\,$\S$ & 7.0 & 13.3 & 6.5 \\

& FsDetView~\cite{FSDetView}\,$\S$ & 10.7 & 24.5 & 6.7 \\

& \cellcolor{black!6}Meta-DETR (Ours)\,$\S$ & \cellcolor{black!6}\textbf{15.4} & \cellcolor{black!6}\textbf{25.0} & \cellcolor{black!6}\textbf{15.8} \\

\midrule[0.68pt]

\multirow{14}{*}{10} & FRCN+ft-full~\cite{FasterRCNN}\,$\ddag$\,$\S$ & 5.5 & 10.0 & 5.5 \\

& Deformable-DETR+ft-full~\cite{DeformableDETR}\,$\S$ & 11.7 & 19.6 & 12.1 \\

& Meta-YOLO~\cite{FewshotReweighting} & 5.6 & 12.3 & 4.6 \\

& Meta\,Det~\cite{metadet} & 7.1 & 14.6 & 6.1 \\

& Meta R-CNN~\cite{metarcnn} & 8.7 & 19.1 & 6.6 \\

& TFA w/ cos~\cite{fsdet}\,$\ddag$\,$\S$ & 9.1 & 17.1 & 8.8 \\

& FSOD~\cite{fsod} & 12.0 & 22.4 & 11.8 \\

& FsDetView~\cite{FSDetView}\,$\S$ & 12.5 & 27.3 & 9.8 \\

& MPSR~\cite{MPSR}\,$\ddag$ & 9.8 & 17.9 & 9.7  \\

& SRR-FSD~\cite{SRR-FSD}\,$\ddag$\,$\uplus$ & 11.3 & 23.0 & 9.8 \\

& CME~\cite{CME}\,$\ddag$ & 15.1 & 24.6 & 16.4 \\

& DCNet~\cite{DenseRelationDistillation}\,$\ddag$\,$\S$ & 12.8 & 23.4 & 11.2 \\

& FSCE~\cite{fsce}\,$\ddag$\,$\S$ & 11.1 & - & 9.8 \\

& \cellcolor{black!6}Meta-DETR (Ours)\,$\S$ & \cellcolor{black!6}\textbf{19.0} & \cellcolor{black!6}\textbf{30.5} & \cellcolor{black!6}\textbf{19.7}\\

\midrule[0.66pt]

\multirow{13}{*}{30} & FRCN+ft-full~\cite{FasterRCNN}\,$\ddag$\,$\S$ & 7.4 & 13.1 & 7.4 \\

& Deformable-DETR+ft-full~\cite{DeformableDETR}\,$\S$ & 16.3 & 27.2 & 16.7 \\

& Meta-YOLO~\cite{FewshotReweighting} & 9.1 & 19.0 & 7.6\\

& Meta\,Det~\cite{metadet} & 11.3 & 21.7 & 8.1\\

& Meta R-CNN~\cite{metarcnn} & 12.4 & 25.3 & 10.8 \\

& TFA w/ cos~\cite{fsdet}\,$\ddag$\,$\S$ & 12.1 & 22.0 & 12.0 \\

& FsDetView~\cite{FSDetView}\,$\S$ & 14.7 & 30.6 & 12.2 \\

& MPSR~\cite{MPSR}\,$\ddag$\ & 14.1 & 25.4 & 14.2 \\

& SRR-FSD~\cite{SRR-FSD}\,$\ddag$\,$\uplus$ & 14.7 & 29.2 & 13.5 \\

& CME~\cite{CME}\,$\ddag$ & 16.9 & 28.0 & 17.8 \\

& DCNet~\cite{DenseRelationDistillation}\,$\ddag$\,$\S$ & 18.6 & 32.6 & 17.5 \\

& FSCE~\cite{fsce}\,$\ddag$\,$\S$ & 15.3 & - & 14.2 \\

& \cellcolor{black!6}Meta-DETR (Ours)\,$\S$ & \cellcolor{black!6}\textbf{22.2} & \cellcolor{black!6}\textbf{35.0} & \cellcolor{black!6}\textbf{22.8} \\

\bottomrule[1.0pt]

\end{tabular}
}
\vspace{+0.35mm}
  \begin{tablenotes}
    \item[1] \textit{"$\ddag$" indicates methods using multi-scale features.}
    \item[2] \textit{"$\S$" indicates results averaged over multiple runs.}
    \item[3] \textit{"$\uplus$" indicates usage of external data.}
  \end{tablenotes}
  
\end{center}
\vspace{-3.0mm}
\end{table}

\subsubsection{MS COCO}

Table\;\ref{tab:Performance_COCO_novel} shows experimental results on MS COCO. It can be seen that, although MS COCO is much more challenging than Pascal VOC with higher complexity like occlusions and large scale variations, Meta-DETR still outperforms all existing methods under all setups by even larger margins. This can be attributed to \textit{(i)} the complete circumvention of even more inaccurate region proposals for novel classes (See Fig.~\ref{fig:fig2_limitation}(a)) caused by the higher complexity of MS COCO, and \textit{(ii)} the effective exploitation of the correlations among more classes in MS COCO. In addition, Meta-DETR performs exceptionally well compared with other region-based methods under the stricter metric AP$_{\rm 0.75}$, which implies that our proposed Meta-DETR can effectively lift the constraint of inaccurate region proposals and produce more accurate few-shot object detection.

\subsection{Ablation Studies}

\begin{table}[t]
\begin{center}
\centering
\caption{Ablation study on region-level detection \textit{vs.} image-level detection}
\vspace{-3.50mm}
\label{tab:ablation1}
\setlength{\tabcolsep}{2.56pt}
\resizebox{0.490\textwidth}{!}{
\begin{tabular}[t]{lcc|ccccc}
\toprule[1.0pt]

\multirow{2}*{Method} & aligned & \multirow{2}*{R/I} & \multicolumn{5}{c}{Novel mAP@0.5} \\

& network & & 1 & 2 & 3 & 5 & 10 \\

\midrule[0.66pt]

FsDetView~\cite{FSDetView} & & R & 24.2 & 35.3 & 42.2 & 49.1 & 57.4 \\

FsDetView\,+\,Deform.\,Trans. & $\checkmark$ & R & \textbf{28.0} & 36.3 & 41.8 & 48.9 & 57.4 \\

Meta-DETR \textit{w/o} CAM & $\checkmark$ & I & 27.2 & \textbf{42.1} & \textbf{50.5} & \textbf{52.9} & \textbf{59.3} \\

\bottomrule[1.0pt]

\end{tabular}
}
\vspace{+0.15mm}
  \begin{tablenotes}
    \item[1] \textit{"R" denotes region-level detection. "I" denotes image-level detection.}
  \end{tablenotes}
\vspace{-1.50mm}
\end{center}
\end{table}

\begin{table}[t]
\begin{center}
\centering
\caption{Ablation study on the impact of Correlational Aggregation Module}
\vspace{-3.66mm}
\label{tab:ablation2}
\setlength{\tabcolsep}{2.75pt}
\resizebox{0.490\textwidth}{!}{
\begin{tabular}[t]{lc|cc|ccccc}
\toprule[1.0pt]

Detection & \multirow{2}*{R/I} & Correlational Aggr. & \multirow{2}{*}{$C$} & \multicolumn{5}{c}{Novel mAP@0.5} \\

Framework &  & Module\, (CAM) & & 1 & 2 & 3 & 5 & 10 \\

\midrule[0.66pt]

\multirow{3}{*}{Meta-DETR} & \multirow{3}{*}{I} &  & 1 & 27.2 & 42.1 & 50.5 & 52.9 & 59.3 \\

& & $\checkmark$ & 1 & 30.3 & 44.0 & 52.1 & 55.7 & \textbf{62.0} \\

& & \cellcolor{black!6}$\checkmark$ & \cellcolor{black!6}5 & \cellcolor{black!6}\textbf{35.1} & \cellcolor{black!6}\textbf{49.0} & \cellcolor{black!6}\textbf{53.2} & \cellcolor{black!6}\textbf{57.4} & \cellcolor{black!6}\textbf{62.0} \\
 
\midrule[0.66pt]

\multirow{2}{*}{FsDetView~\cite{FSDetView}} & \multirow{2}{*}{R} &  & 1 & 24.2 & 35.3 & 42.2 & 49.1 & 57.4 \\

& &  $\checkmark$ & 5 & \textbf{30.1} & \textbf{41.1} & \textbf{45.2} & \textbf{51.4} & \textbf{57.5} \\

\bottomrule[1.0pt]

\end{tabular}
}
\vspace{+0.15mm}
  \begin{tablenotes}
    \item[1] \textit{"R" denotes region-level detection. "I" denotes image-level detection.}
    \vspace{+0.6777mm}
    \item[2] \textit{"C" denotes the number of support classes to aggregate simultaneously, which can only be 1 without the proposed Correlational Aggregation Module (CAM).}
  \end{tablenotes}
\vspace{-3.0mm}
\end{center}
\end{table}

We conduct comprehensive ablation studies to verify the effectiveness of our design choices. Experimental results are averaged over 10 runs with different randomly sampled support datasets on the first class split of Pascal VOC.

\vspace{+0.88mm}
\smallskip \noindent
\textbf{\textit{Region-Level Detection vs. Image-Level Detection.\;\;}}
From Table\;\ref{tab:Performance_VOC_novel} and Table\;\ref{tab:Performance_COCO_novel}, we can find that fine-tuning Deformable DETR (Deformable-DETR+ft-full) generally outperforms fine-tuning Faster R-CNN (FRCN+ft-full), especially in the MS COCO dataset, where it is much harder to obtain accurate region proposals for novel classes due to higher complexity (see Fig.\;\ref{fig:fig2_limitation}(a)). This observation aligns well with our insight that region-based detection frameworks tend to suffer from inaccurate regional proposals for novel classes. To further verify the superiority of image-level few-shot object detection, we adopt FsDetView~\cite{FSDetView}, a state-of-the-art meta-learning-based few-shot detector built on top of Faster R-CNN, as a solid baseline to compare with our method. For a fair comparison, we add deformable transformers to FsDetView (denoted as FsDetView\,+\,Deform.\,Trans.) to rule out the performance difference brought by the transformer architecture. Furthermore, we replace our proposed CAM in Meta-DETR with the feature aggregation module proposed in FsDetView (denoted as Meta-DETR\,\textit{w/o}\,CAM). As shown in Table\;\ref{tab:ablation1}, even with aligned network architecture and aggregation scheme, Meta-DETR\,\textit{w/o}\,CAM still outperforms FsDetView\,+\,Deform.\,Trans. under most setups. The results validate the superiority of solving few-shot object detection at image level.

\begin{figure*}[t!] 
\begin{center}
   \includegraphics[width=1.0\linewidth]{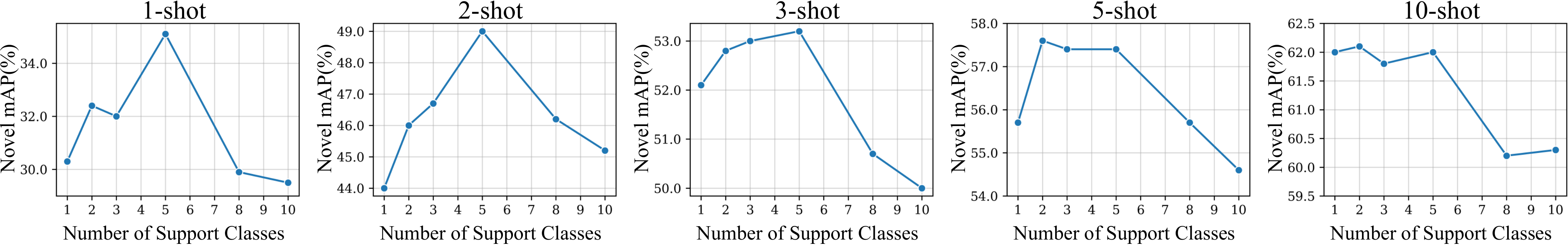}
\end{center}
\vspace{-3.80mm}
   \caption{
   Ablation study on the number of support classes for simultaneous correlational aggregation under different few-shot setups. Results are averaged over 10 repeated runs on Pascal VOC class split 1.
   }
   \label{fig:fig5_num_supp_class}
\vspace{-1.0mm}
\end{figure*}

\begin{figure}[t]
\begin{center}
   \includegraphics[width=0.95\linewidth]{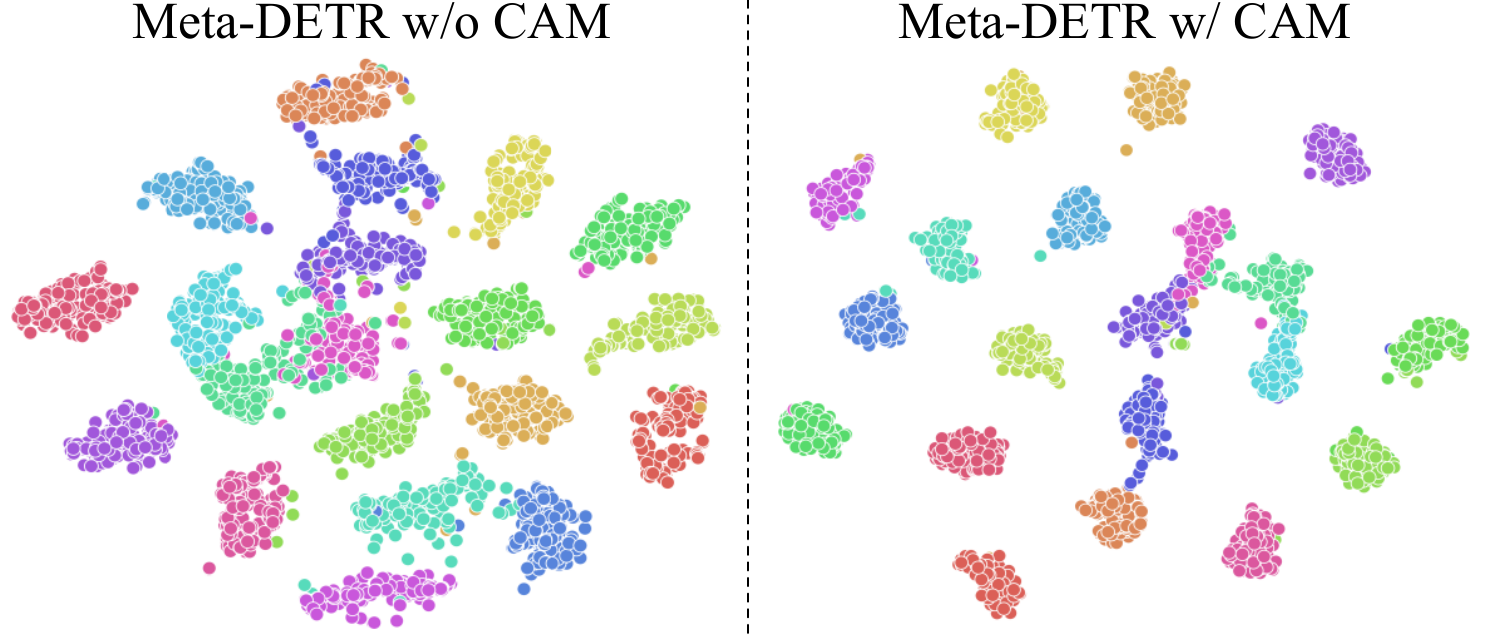}
\end{center}
\vspace{-3.50mm}
   \caption{
   t-SNE visualization of objects learned in the feature space with and without our designed Correlational Aggregation Module. Results are obtained on Pascal VOC class split 1 under the 2-shot setup.}
\label{fig:fig6_tsne}
\end{figure}

\begin{table}[t]
\begin{center}
\centering
\caption{Confusion matrices of similar class pairs predicted with \\ and without the proposed Correlational Aggregation Module}
\label{tab:confusion_matrix}
\vspace{-2.0mm}
\centering
\begin{center}
\includegraphics[width=0.850\linewidth]{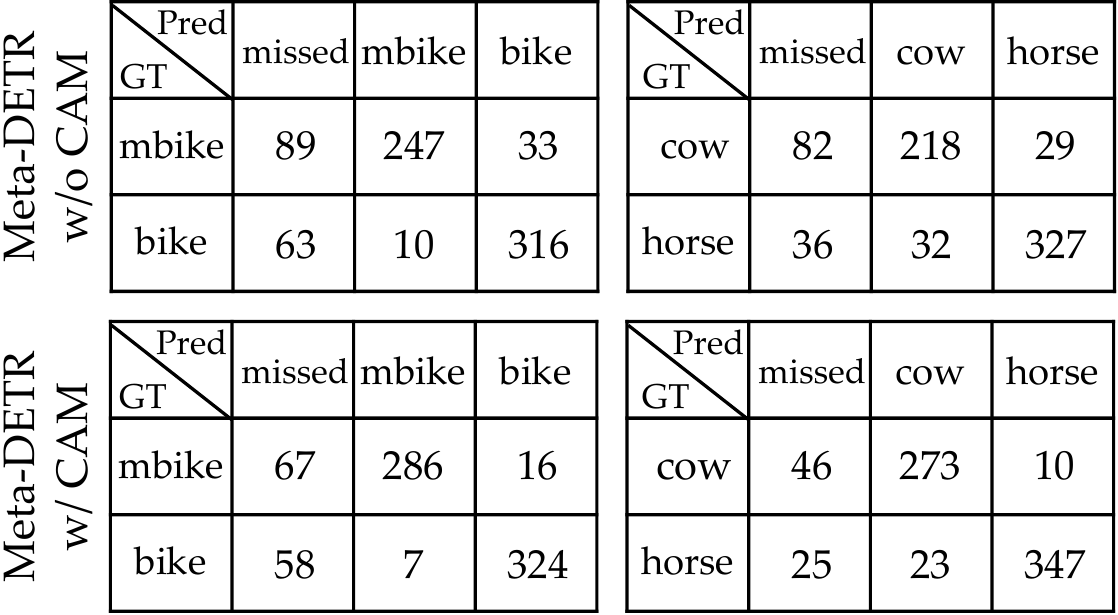}
\end{center}
\vspace{+0.5mm}
  \begin{tablenotes}
    \item[1] \textit{Results obtained on Pascal VOC class split 1 under the 2-shot setup.}
    \vspace{+0.25mm}
    \item[1] \textit{"GT" denotes ground truth label; "Pred" denotes predicted label.}
  \end{tablenotes}
\vspace{-2.50mm}
\end{center}
\end{table}

\vspace{+0.88mm}
\smallskip \noindent
\textit{\textbf{Impact of Correlational Aggregation Module (CAM).\;\;}}
As shown in Table\;\ref{tab:ablation2}, when incorporating CAM into our model, even if we keep the number of support classes for simultaneous aggregation ($C$) as 1, CAM can still boost few-shot detection performance under all settings. This demonstrates CAM's strong capacity in aggregating query and support information even without the leverage of inter-class correlation. When multiple support classes are available ($C\geq$2), CAM can further exploit their inter-class correlation to boost few-shot detection performance under lower-shot ($\leq$5) settings, especially under 1-shot (+\,4.8\%\;mAP) and 2-shot (+\,5.0\%\;mAP), which shows the benefit of inter-class correlational meta-learning. No clear performance gain is observed for 10-shot, which implies that, when more training samples are available, the detector can already recognize novel classes and differentiate them from similar classes without explicitly modeling the inter-class correlation. We also apply our designed CAM to the commonly used region-based meta-detector FsDetView~\cite{FSDetView} and report the results in Table\;\ref{tab:ablation2}. Its steady performance gain demonstrates that CAM and the proposed inter-class correaltional meta-learning strategy can also benefit region-level few-shot object detection.

To understand how CAM functions to improve detection accuracy, we visualize the objects from different classes in the feature space learned with and without the proposed CAM with t-SNE~\cite{tsne}. As shown in Fig.\;\ref{fig:fig6_tsne}, with CAM included to perform inter-class correlational meta-learning, object classes are better separated from each other, which affirms our motivation of leveraging inter-class correlation to reduce misclassification among similar classes. To further verify our claim that CAM effectively reduces misclassification among similar classes, we select two pairs of similar classes (\textit{motorbike\;vs.\;bike} and \textit{cow\;vs.\;horse}) and plot their confusion matrices in Table\;\ref{tab:confusion_matrix}. We can observe that CAM indeed reduces the misclassification by large margins with the exploitation of inter-class correlation. We also observe fewer missed predictions, which shows that the effective leverage of inter-class correlations also facilitates generalization to detect previously missed cases.

\smallskip
\vspace{+0.3mm} \noindent
\textit{\textbf{Number of Classes for Correlational Aggregation.\;\;}}
Meta-DETR receives a fixed number of support classes ($C$) and simultaneously aggregates them with query features to capture the inter-class correlation among different support classes. With $C\geq2$, Meta-DETR exploits the inter-class correlation among different classes. Fig.\;\ref{fig:fig5_num_supp_class} investigates the impact of the number of support classes for aggregation. As the number of support classes $C$ increases from 1 to 10, the lower-shot ($\leq$5) detection performance first improves and then drops, while 10-shot performance first saturates and then drops. This validates the effectiveness of leveraging inter-class correlation under lower-shot ($\leq$5) settings. The performance gain is considerable under extremely low-shots like 1-shot and 2-shot, indicating that it is highly beneficial to explore inter-class correlation when training samples are too scarce to model a novel class and differentiate it with other classes. We conjecture that the performance drop with a large number of support classes ($\geq$8) for correlational aggregation is due to the model's limited capacity to differentiate too many support classes at one go. Based on the results, we set our method's number of support classes $C$ as 5 in all other experiments unless otherwise stated.

\begin{table}[t]
\vspace{-0.75mm}
\begin{center}
\centering
\caption{Ablation study on the design choices of the attention mechanism \\ in the proposed Correlational Aggregation Module}
\label{tab:cam_ablation}
\vspace{-5.00mm}
\setlength{\tabcolsep}{4.0pt}
\resizebox{0.486\textwidth}{!}{
\begin{tabular}[t]{ccc|ccccc}
\toprule[1.00pt]

\textit{(a)} Apply & \textit{(b)} Query & \textit{(c)} Modeling & \multicolumn{5}{c}{Novel mAP@0.5} \\

 Sigmoid & Multiplication & Background & 1 & 2 & 3 & 5 & 10 \\

\midrule[0.68pt]

  &  &  & 29.8 & 44.8 & 51.2 & 54.8 & 59.6 \\

  & & $\checkmark$ & 31.2 & 46.1 & 52.5 & 56.2 & 61.5 \\

 $\checkmark$ & $\checkmark$ & & 32.6 &  45.6  & 51.3 &  56.1 &  60.9 \\

 \cellcolor{black!6}$\checkmark$ & \cellcolor{black!6}$\checkmark$ & \cellcolor{black!6}$\checkmark$ & \cellcolor{black!6}\textbf{35.1} & \cellcolor{black!6}\textbf{49.0} & \cellcolor{black!6}\textbf{53.2} & \cellcolor{black!6}\textbf{57.4} & \cellcolor{black!6}\textbf{62.0} \\

\bottomrule[1.00pt]

\end{tabular}
}
\vspace{-1.5mm}
\end{center}
\end{table}

\begin{figure*}[t] 
\begin{center}
   \includegraphics[width=1.0\linewidth]{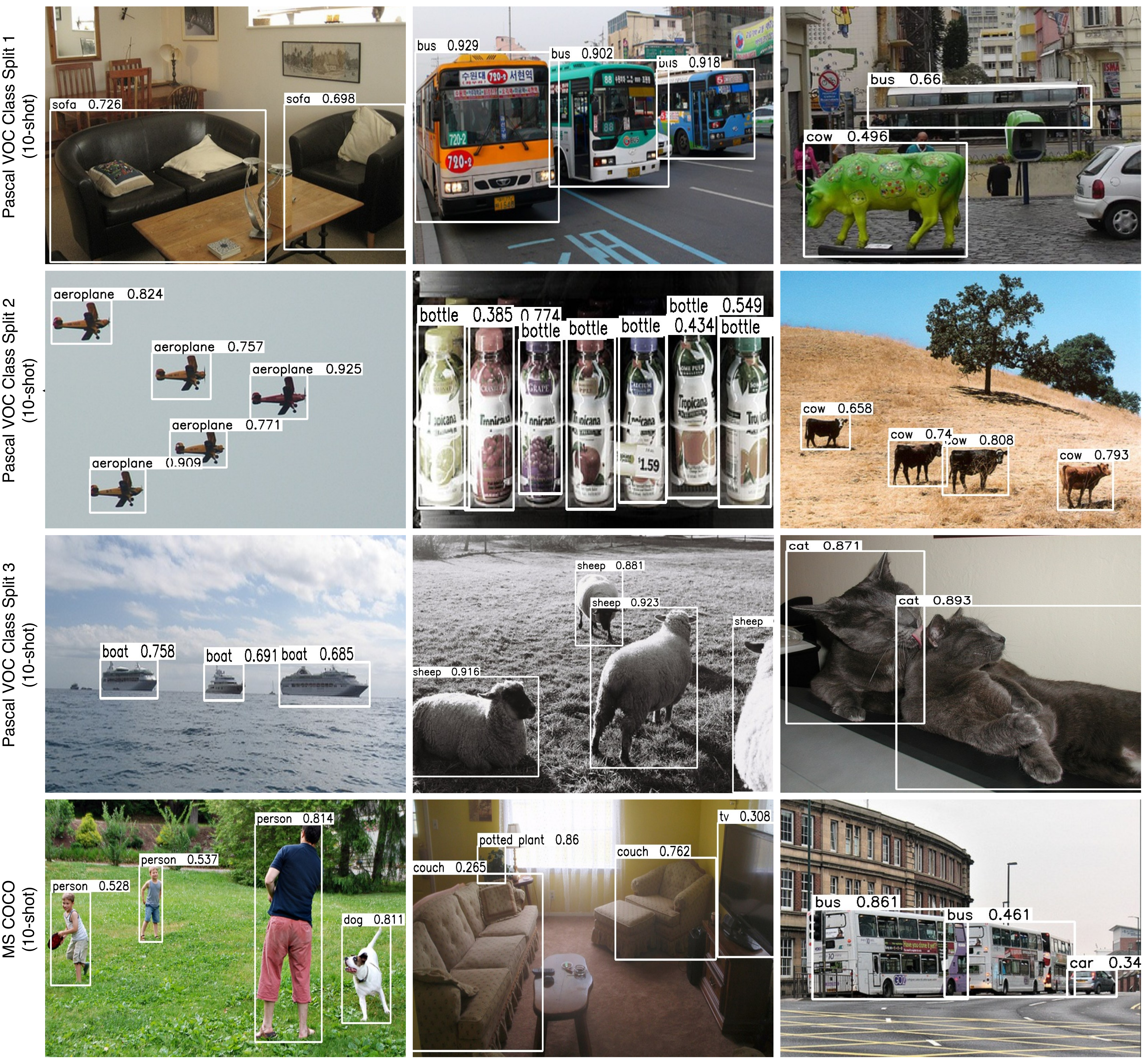}
\end{center}
\vspace*{-4.486mm}
\caption{Visualization of Meta-DETR's 10-shot object detection results on various data setups. For simplicity, only detections of novel-class objects are illustrated. The qualitative experimental results show that Meta-DETR can detect novel objects effectively with very constrained training samples.}
\label{fig:qualitative_exp_result}
\vspace{+1.25mm}
\end{figure*}

\begin{table}[t]
\begin{center}
\centering
\caption{Ablation study on early aggregation \textit{vs.} late aggregation}
\label{tab:cam_loc}
\vspace{-4.0mm}
\setlength{\tabcolsep}{7.25pt}
\resizebox{0.4486\textwidth}{!}{
\begin{tabular}[t]{c|ccccc}
\toprule[1.0pt]

CAM's Location & \multicolumn{5}{c}{Novel mAP@0.5} \\

@ Encoder Layers & 1 & 2 & 3 & 5 & 10 \\

\midrule[0.56pt]

\cellcolor{black!6}1 & \cellcolor{black!6}\textbf{35.1} & \cellcolor{black!6}\textbf{49.0} & \cellcolor{black!6}\textbf{53.2} & \cellcolor{black!6}\textbf{57.4} & \cellcolor{black!6}\textbf{62.0} \\

3 &  27.1 & 42.9 & 50.6 & 54.0 & 59.2 \\

6 & 15.2 & 31.5 & 37.7 & 50.3 & 53.4 \\

\bottomrule[1.0pt]

\end{tabular}
}
\vspace{-1mm}
\end{center}
\end{table}

\begin{figure*}[t] 
\begin{center}
   \includegraphics[width=0.99\linewidth]{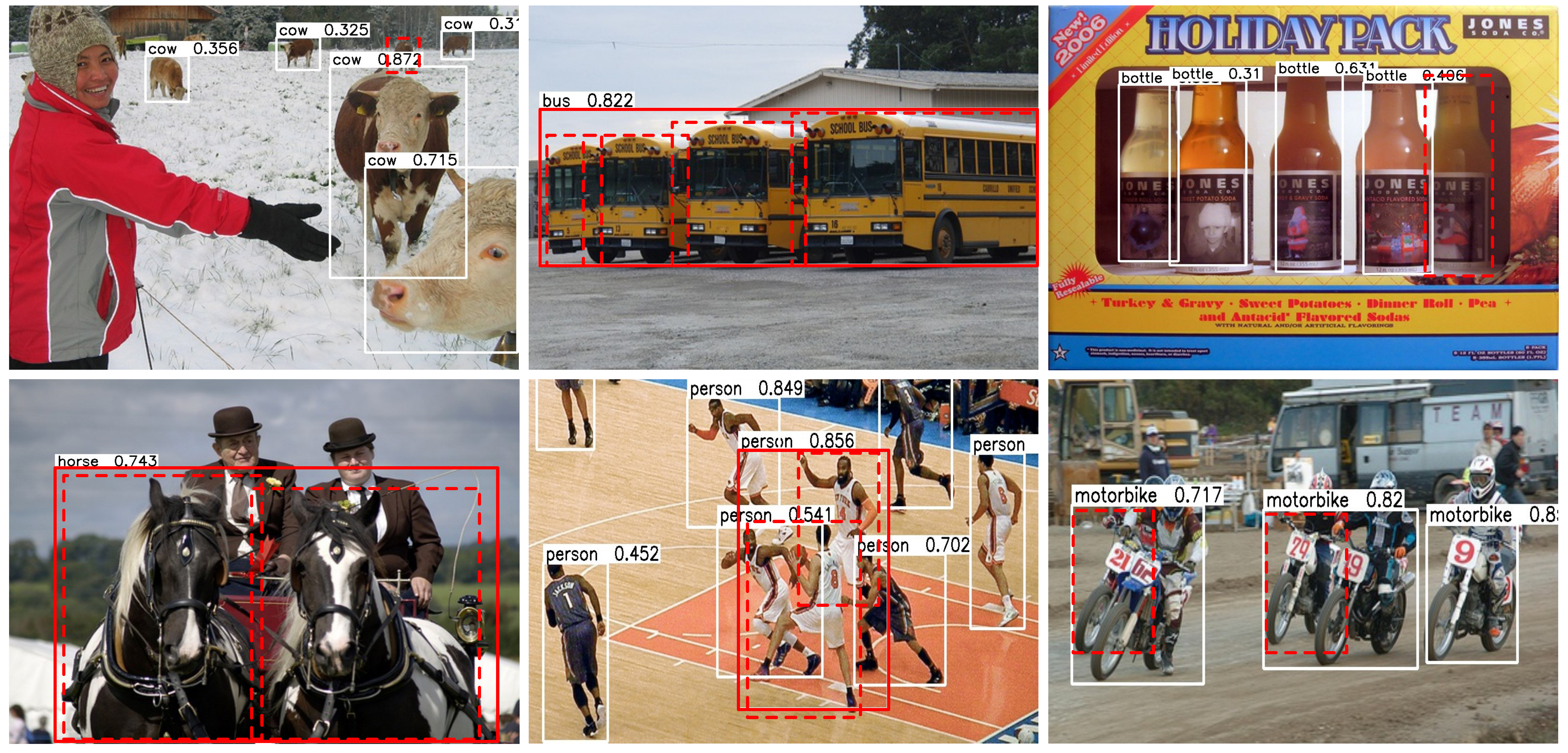}
\end{center}
\vspace*{-4.86mm}
\caption{Visualization of some failure cases of Meta-DETR's 10-shot object detection results. For simplicity, only detections of novel-class objects are illustrated. White boxes indicate true positives. Red solid boxes indicate false positives. Red dashed boxes indicate false negatives.}
\label{fig:failure_cases}
\vspace{+1.5mm}
\end{figure*}

\begin{table*}[t]
\begin{center}
\centering
\caption{Few-shot object detection and instance segmentation performance on COCO for novel classes}
\vspace{-5.0mm}
\label{tab:Performance_COCO_nove_instance_segmentation}
\setlength{\tabcolsep}{4.486pt}
\resizebox{0.99\textwidth}{!}{
\begin{tabular}[t]{c|l|cccccc|cccccc}
\toprule[1.1pt]

\multirow{2}*{Shot} & \multirow{2}*{Method} & \multicolumn{6}{c|}{Box} & \multicolumn{6}{c}{Mask} \\

\cline{3-14}

& & AP$_{\rm 0.5:0.95}$ & AP$_{\rm 0.5}$ & AP$_{\rm 0.75}$ & AP$_{\rm S}$ & AP$_{\rm M}$ & AP$_{\rm L}$ & AP$_{\rm 0.5:0.95}$ & AP$_{\rm 0.5}$ & AP$_{\rm 0.75}$ & AP$_{\rm S}$ & AP$_{\rm M}$ & AP$_{\rm L}$ \\

\midrule

\multirow{3}*{5} & Mask-RCNN+ft-full~\cite{MaskRCNN} & 1.3 & 3.0 & 1.1 & 0.3 & 1.1 & 2.4 & 1.3 & 2.7 & 1.1 & 0.3 & 0.6 & 2.2 \\

& Meta R-CNN~\cite{metarcnn} & 3.5 & 9.9 & 1.2 & 1.2 & 3.9 & 5.8 & 2.8 & 6.9 & 1.7 & 0.3 & 2.3 & 4.7 \\

& \cellcolor{black!6}Meta-DETR (Ours) & \cellcolor{black!6}\textbf{15.3} & \cellcolor{black!6}\textbf{24.9} & \cellcolor{black!6}\textbf{15.4} & \cellcolor{black!6}\textbf{1.5} & \cellcolor{black!6}\textbf{12.8} & \cellcolor{black!6}\textbf{26.0} & \cellcolor{black!6}\textbf{8.1} & \cellcolor{black!6}\textbf{16.8} & \cellcolor{black!6}\textbf{7.1} & \cellcolor{black!6}\textbf{0.9} & \cellcolor{black!6}\textbf{5.6} & \cellcolor{black!6}\textbf{13.7} \\

\midrule

\multirow{3}*{10} & Mask-RCNN+ft-full~\cite{MaskRCNN} & 2.5 & 5.7 & 1.9 & 2.0 & 2.7 & 3.9 & 1.9 & 4.7 & 1.3 & 0.2 & 1.4 & 3.2 \\

& Meta R-CNN~\cite{metarcnn} & 5.6 & 14.2 & 3.0 & 2.0 & 6.6 & 8.8 & 4.4 & 10.6 & 3.3 & 0.5 & 3.6 & 7.2 \\

& \cellcolor{black!6}Meta-DETR (Ours) & \cellcolor{black!6}\textbf{19.8} & \cellcolor{black!6}\textbf{31.3} & \cellcolor{black!6}\textbf{20.4} & \cellcolor{black!6}\textbf{4.5} & \cellcolor{black!6}\textbf{17.4} & \cellcolor{black!6}\textbf{30.5} & \cellcolor{black!6}\textbf{10.1} & \cellcolor{black!6}\textbf{20.8} & \cellcolor{black!6}\textbf{8.7} & \cellcolor{black!6}\textbf{1.7} & \cellcolor{black!6}\textbf{7.6} & \cellcolor{black!6}\textbf{15.8} \\

\bottomrule[1.1pt]

\end{tabular}
}

\end{center}
\vspace{+1.0mm}
\end{table*}

\smallskip
\vspace{+0.3mm} \noindent
\textit{\textbf{Design Choices for Correlational Aggregation Module (CAM).\;\;}}
The proposed CAM's attention mechanism differs from the original DETR attention in three aspects: \textit{(a)} applying a sigmoid function to attention's \textit{Value} in feature matching, \textit{(b)} multiplying attention's output with attention's \textit{Query} in feature matching, and \textit{(c)} explicitly modelling a prototype for the `background' class. Among them, \textit{(a)} and \textit{(b)} are designed as a whole with \textit{(a)} for generating `filters' to remove query features that are irrelevant to the given support classes and \textit{(b)} for applying the learned `filters' to the query image features. And \textit{(c)} enables Meta-DETR to better handle the `no match' scenario where the query features do not match any of the support classes.  We present ablation experiments in Table\;\ref{tab:cam_ablation} that verify the effectiveness of the above three modifications.

\smallskip
\vspace{+0.5mm} \noindent
\textit{\textbf{Early Aggregation vs. Late Aggregation.\;\;\;}}
The proposed CAM replaces one encoder layer in the transformer. As shown in Fig.~\ref{fig:fig3_architecture}, we place CAM ahead of the transformer encoder (as the first layer of the encoder). Table~\ref{tab:cam_loc} studies the impact of the location of CAM in the transformer encoder. As shown, it is preferable to place CAM at the beginning stage of the transformer encoder for early aggregation, which also suggests the importance of learning a deep class-agnostic predictor.

\subsection{Qualitative Results}

In Fig.\;\ref{fig:qualitative_exp_result}, we provide qualitative visualization of Meta-DETR's 10-shot object detection results on several sample images from their respective data setups. Note that we show the detection of novel classes only since the focus of few-shot object detection is to detect objects of novel classes. We show detection results with confidence scores higher than 0.25 to filter out low-confidence predictions. It can be observed that the proposed Meta-DETR is able to detect novel objects effectively even with very limited training samples.


\subsection{Failure Cases and Future Directions}

Fig.\;\ref{fig:failure_cases} illustrates typical failure cases of the proposed Meta-DETR. The most typical failure cases happen while multiple instances of novel objects are heavily clustered, largely due to the lack of supervision in such cases and the lack of a mechanism to discriminate objects' boundaries. Other typical failure cases include difficulty in detecting small objects as well as false negatives with less salient objects, which are also applicable in general object detectors.

Although the current few-shot object detection performance is still far from perfect, our proposed Meta-DETR establishes a new few-shot object detection paradigm that is conceptually simple with room for improvement. In our future work, we will investigate new mechanisms that can highlight object boundaries and thus help avoid some failure case as illustrated in Fig.\;\ref{fig:failure_cases}. Besides, since we only explore single-scale features throughout all experiments, an interesting and promising direction is to exploit multi-scale features in meta-learning-based few-shot object detection. By properly designing a dual-scale-selection strategy for both query and support, we expect it can further improve the performance of few-shot object detection, especially on small objects.

\begin{figure}[t]
\begin{center}
   \includegraphics[width=1.0\linewidth]{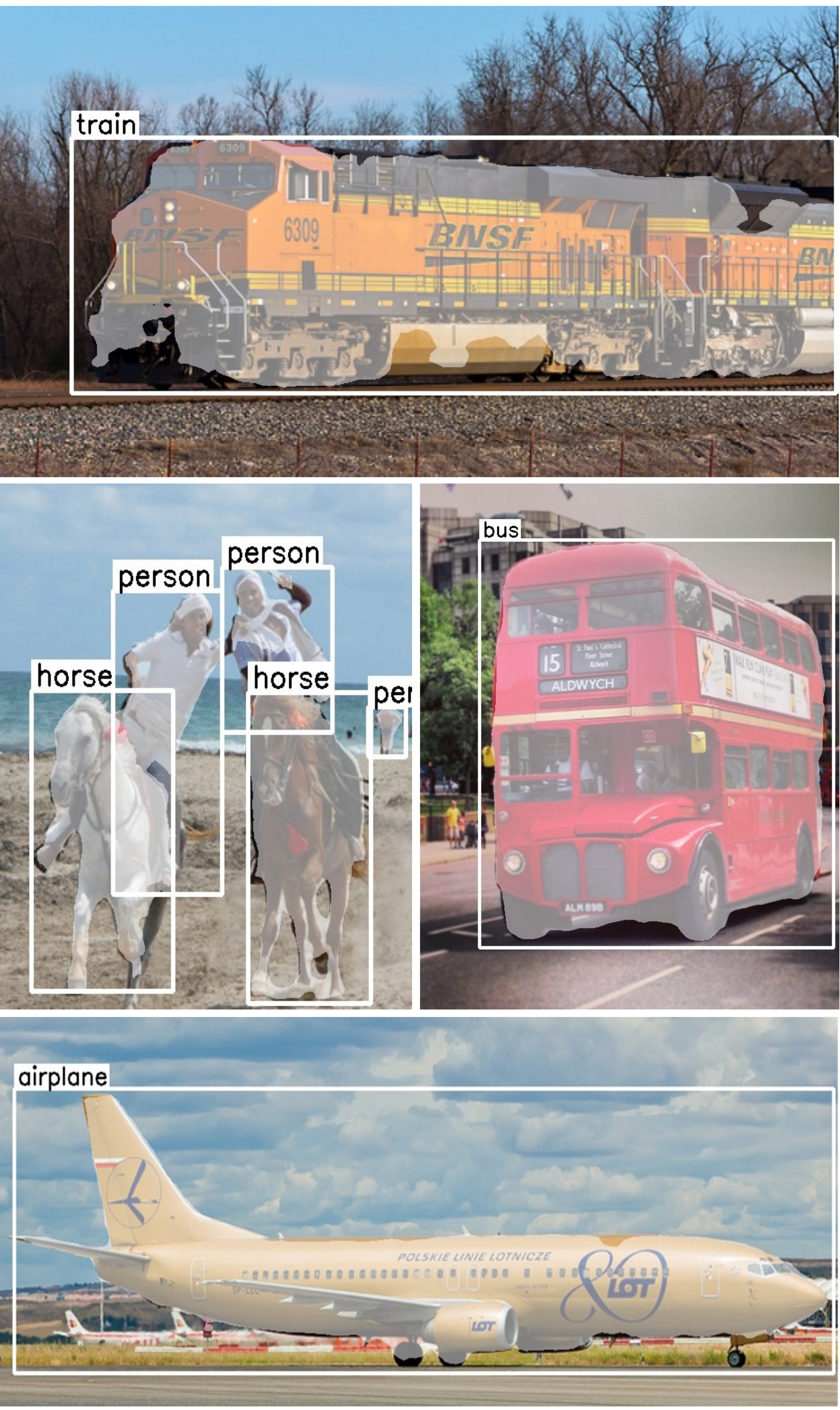}
\end{center}
\vspace{-4.86mm}
   \caption{
   Visualization of Meta-DETR's 10-shot instance segmentation results on MS COCO. For simplicity, only segmentations of novel-class objects are illustrated.}
\label{fig:qualitative_exp_result_instance_segmentation}
\vspace{+2mm}
\end{figure}

\subsection{Extension to Few-Shot Instance Segmentation}

The proposed Meta-DETR adopts a meta-learning framework which is generic and can be adapted to other downstream vision tasks beyond object detection. We validate this feature by examining how it can be extended to perform instance segmentation with simple modifications.

As described in~\cite{DETR}, the original DETR can be extended to perform instance segmentation by adding a mask head on top of the decoder outputs. We similarly introduce an additional mask head over Meta-DETR to predict objects' masks for few-shot instance segmentation. The additional mask head takes the output of the transformer decoder and encoded image features as input and predicts a binary mask for each object query. It also follows the designed inter-class correlational meta-learning strategy for better generalization. To train Meta-DETR to perform few-shot instance segmentation, we first train it on the previously mentioned few-shot object detection tasks, and then freeze all the weights and train only the additional mask head for instance segmentation.

\smallskip
\vspace{+1.5mm} \noindent
\textit{\textbf{Experimental Results.\;\;}} We conduct experiments for few-shot instance segmentation on MS COCO under 5-shot and 10-shot setups. Similarly, the 20 classes shared with Pascal VOC are chosen as novel classes, and the remaining 60 classes are set as base classes. Note that AP for instance segmentation is evaluated with mask IoU. As shown in Table~\ref{tab:Performance_COCO_nove_instance_segmentation}, Meta-DETR outperforms compared methods by large margins. The results demonstrate the superiority and universality of our Meta-DETR, which can extend to other instance-level few-shot learning tasks. Note that the compared Meta R-CNN~\cite{metarcnn} adopts region-level prediction together with the conventional class-by-class meta-learning via feature reweighting. The comparison between Meta R-CNN~\cite{metarcnn} and our proposed Meta-DETR verifies that the combination of the image-level prediction and the exploitation of inter-class correlation via correlational meta-learning can effectively benefit other instance-level few-shot learning tasks like few-shot instance segmentation. We also provide qualitative results for instance segmentation in Fig.~\ref{fig:qualitative_exp_result_instance_segmentation}.

\section{Conclusion}

This paper presents a new few-shot object detection framework, namely Meta-DETR. The proposed framework achieves \textit{(i)} pure image-level prediction, which lifts the constraints caused by novel classes' inaccurate region proposals, and \textit{(ii)} effective exploitation of categorical correlation via a inter-class correlational meta-learning strategy, which reduces misclassification and enhances generalization among similar or related classes. Despite its simplicity, our method achieves state-of-the-art performance over multiple few-shot object detection setups, outperforming prior works by large margins. It can also be easily extended to other instance-level few-shot learning tasks. We hope this work can offer good insights and inspire further researches in few-shot object detection and other related topics.


%



\ifCLASSOPTIONcompsoc
  
  \section*{Acknowledgment}
\else
  \section*{Acknowledgment}
\fi

This research is supported by the Ministry of Education, Singapore, under its Academic Research Fund Tier 1 (RG94/20).

\ifCLASSOPTIONcaptionsoff
  \newpage
\fi


\newpage
\bibliographystyle{IEEEtran}
\bibliography{bib.bib}

%

\newpage 

\begin{IEEEbiography}[{\includegraphics[width=1in,height=1.25in,clip,keepaspectratio]{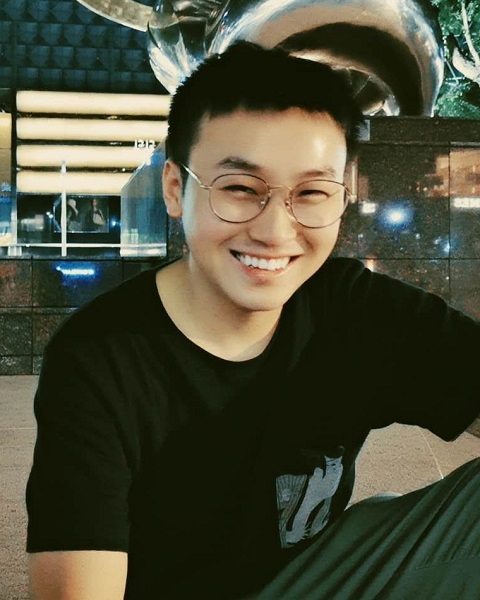}}]{Gongjie Zhang}
is currently working toward the Ph.D. degree in the School of Computer Science and Engineering, Nanyang Technological University, Singapore, under the supervision of Dr.\,Shijian Lu. He received his B.Eng. degree in electronic and information engineering in 2018 from Northeastern University, Shenyang, China. He has  published multiple journal and conference papers in the field of computer vision. He has also served as reviewer for several top journals and conferences such as T-PAMI, T-IP, CVPR, ICCV, and ACM\,MM. His research interests mainly include computer vision, object detection, few-shot learning, and meta-learning.
\end{IEEEbiography}

\begin{IEEEbiography}[{\includegraphics[width=1in,height=1.25in,clip,keepaspectratio]{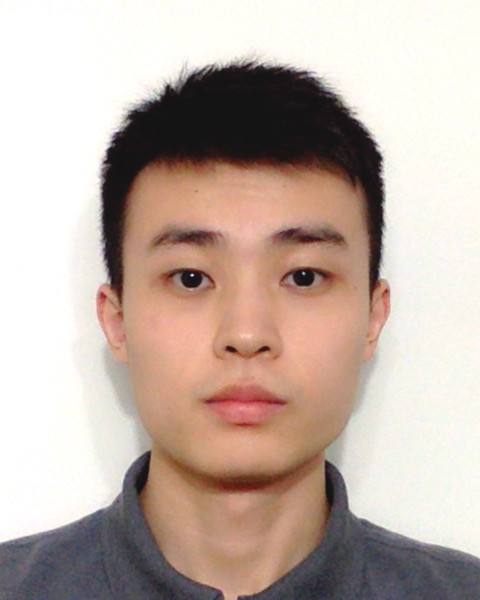}}]{Zhipeng Luo}
is currently a Ph.D. student with the School of Computer Science and Engineering, Nanyang Technological University, Singapore, under the supervision of Dr.\,Shijian Lu. He received his B.Eng. degree in mechanical engineering in 2015 and M.Sc. degree in computing in 2018 from National University of Singapore. He has  published multiple top conference papers in the field of computer vision. His research interests include computer vision, object detection, and object tracking.
\end{IEEEbiography}

\begin{IEEEbiography}[{\includegraphics[width=1in,height=1.25in,clip,keepaspectratio]{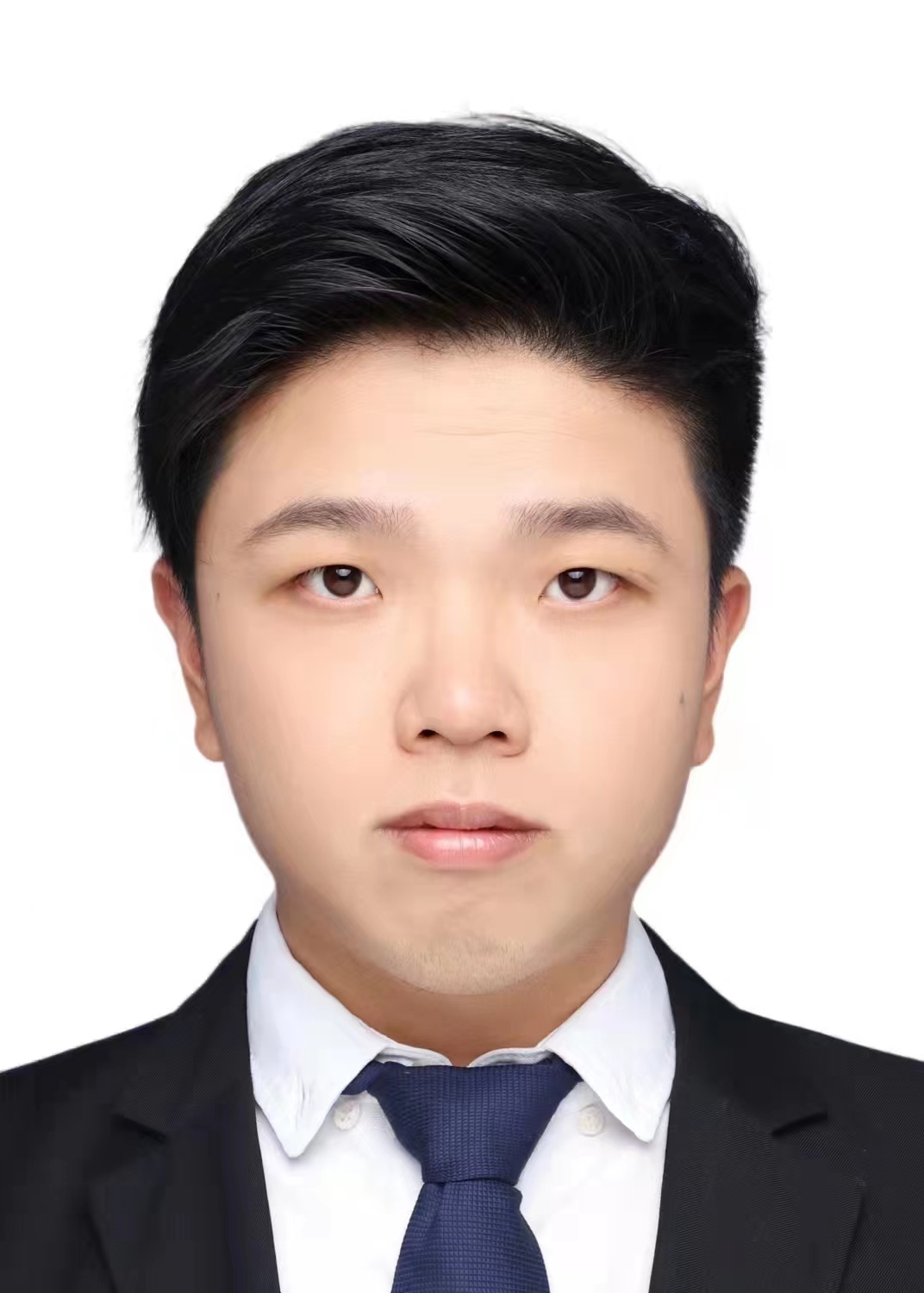}}]{Kaiwen Cui}
is currently working toward the Ph.D. degree in the School of Computer Science and Engineering, Nanyang Technological University, Singapore, under the supervision of Dr.\,Shijian Lu. He received his B.Eng. degree in 2016 and M.Sc. degree in 2017, both in electrical and electronic engineering from National University of Singapore. He has  published multiple top conference papers in the field of computer vision. His research interests mainly include computer vision and data-limited image generation.
\end{IEEEbiography}

\newpage

\begin{IEEEbiography}[{\includegraphics[width=1in,height=1.25in,clip,keepaspectratio]{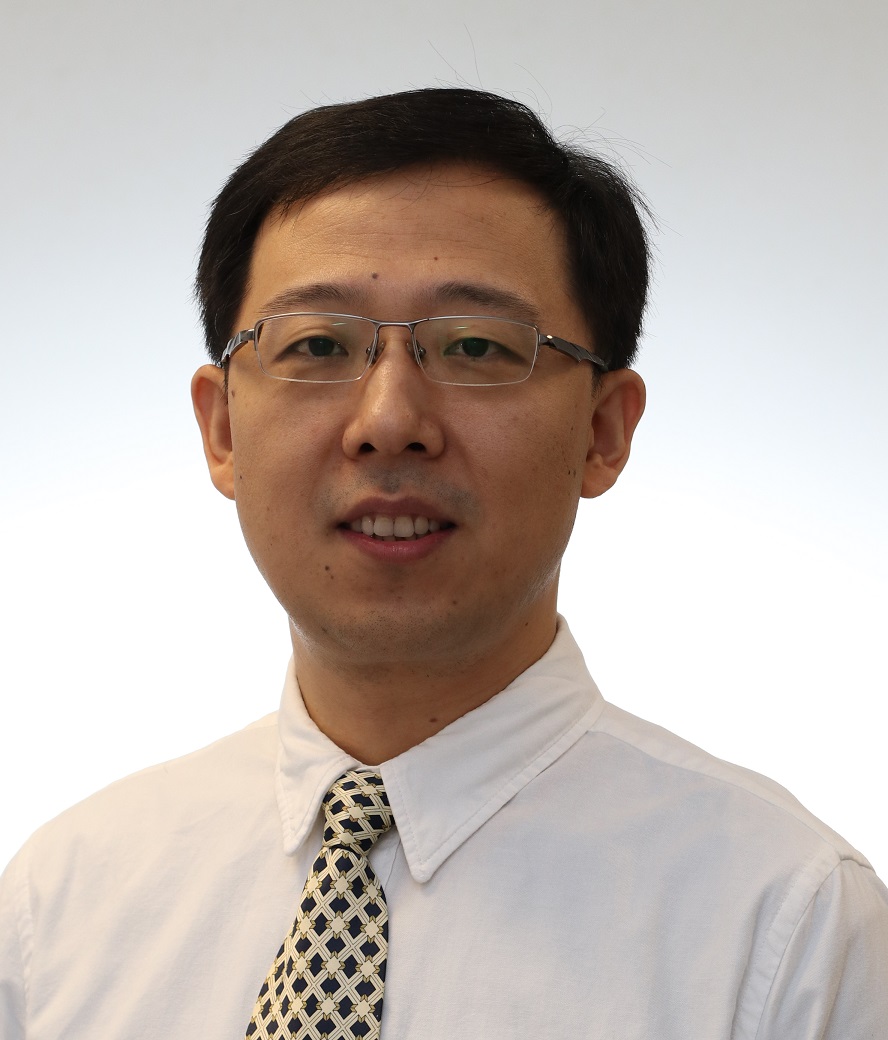}}]{Shijian Lu}
received his Ph.D. in electrical and computer engineering from National University of Singapore. He is an Associate Professor with the School of Computer Science and Engineering, Nanyang Technological University, Singapore. His major research interests include image and video analytics, visual intelligence, and machine learning. He has published more than 100 international refereed journal and conference papers and co-authored over 10 patents in these research areas. He is currently an Associate Editor for the journal Pattern Recognition (PR). He has also served in the program committee of a number of conferences, \textit{e.g.}, the Area Chair of the International Conference on Document Analysis and Recognition (ICDAR) 2017 and 2019, the Senior Program Committee of the International Joint Conferences on Artificial Intelligence (IJCAI) 2018 and 2019, \textit{etc}.
\end{IEEEbiography}

\begin{IEEEbiography}[{\includegraphics[width=1in,height=1.25in,clip,keepaspectratio]{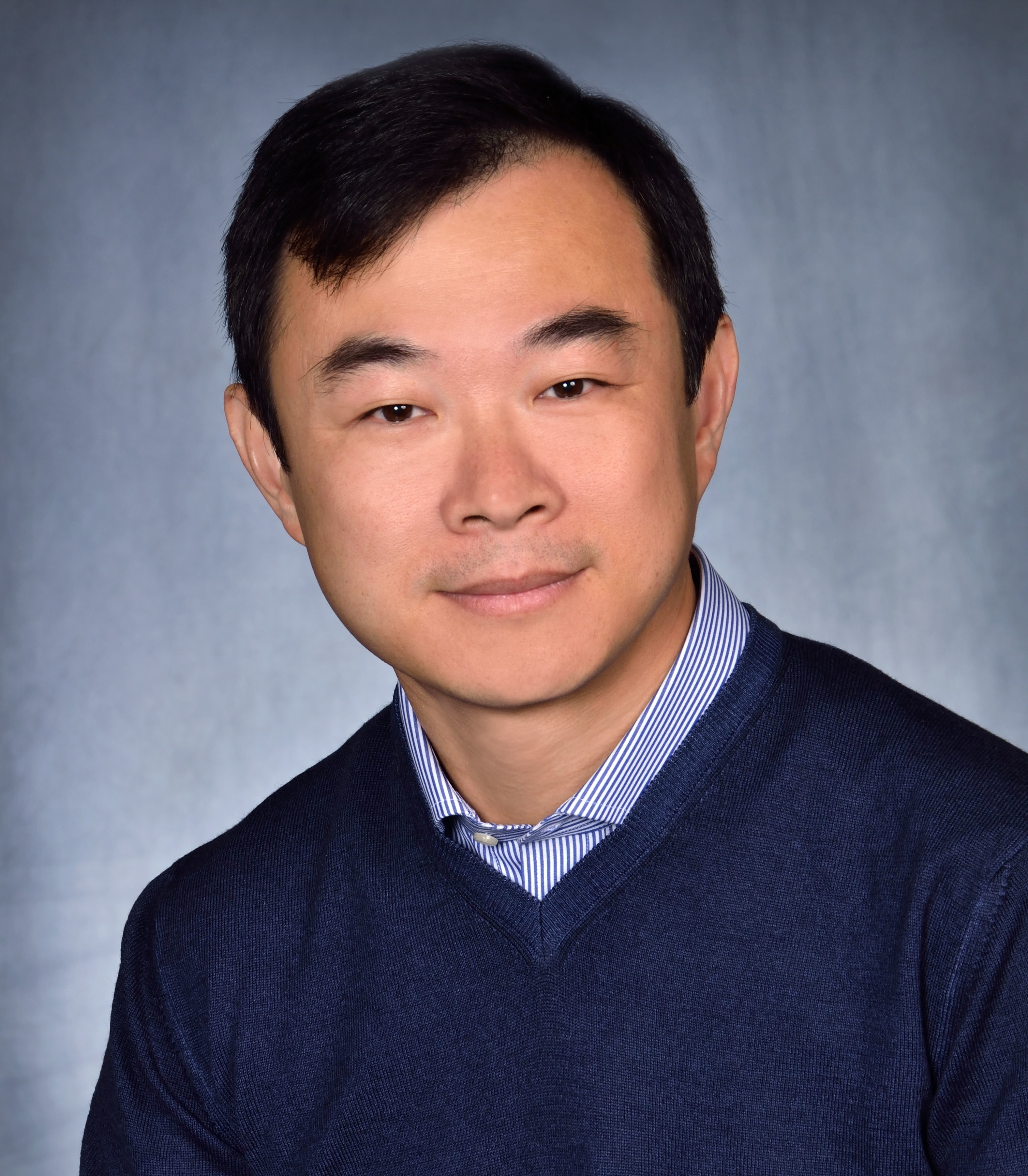}}]{Eric P. Xing}
(Fellow, IEEE) received the Ph.D. degree in molecular biology from Rutgers University, New Brunswick, NJ, USA, in 1999, and the Ph.D. degree in computer science from the University of California at Berkeley, Berkeley, CA, USA, in 2004. He is currently a Professor of machine learning with the School of Computer Science and the Director of the CMU Center for Machine Learning and Health, Carnegie Mellon University, Pittsburgh, PA, USA. His principal research interests lie in the development of machine learning and statistical methodology, especially for solving problems involving automated learning, reasoning, and decision-making in high-dimensional, multimodal, and dynamic possible worlds in social and biological systems. Dr.\,Xing is a member of the DARPA Information Science and Technology (ISAT) Advisory Group and the Program Chair of the International Conference on Machine Learning (ICML) 2014. He is also an Associate Editor of The Annals of Applied Statistics (AOAS), the Journal of American Statistical Association (JASA), the IEEE Transactions on Pattern Analysis and Machine Intelligence (T-PAMI), and PLOS Computational Biology and an Action Editor of the Machine Learning Journal (MLJ) and the Journal of Machine Learning Research (JMLR).
\end{IEEEbiography}





\end{document}